\documentclass[journal]{IEEEtran}
\usepackage{graphicx}
\usepackage{subfigure}
\usepackage{cite}
\usepackage{amsthm}
\usepackage{amsmath}
\usepackage{algorithm} 
\usepackage{algorithmic}
\usepackage{rotating}
\usepackage{color}

\makeatletter

\newcommand{\Rmnum}[1]{\expandafter\@slowromancap\romannumeral #1@}
\makeatother
 
\begin{document}

\title{An Efficient and Accurate Rough Set for Feature Selection, Classification and Knowledge Representation}

\author{Shuyin~Xia, 
		Xinyu~Bai, 
		Guoyin~Wang*, 
		Deyu~Meng, 
		Xinbo~Gao,
		Zizhong~Chen, 
		Elisabeth~Giem
\thanks{S. Xia, X. Bai, G. Wang \& X. Gao are with the Chongqing Key Laboratory of Computational Intelligence, Chongqing University of Telecommunications and Posts, Chongqing, 400065, China. E-mail: xiasy@cqupt.edu.cn, 1025476698@qq.com, wanggy@cqupt.edu.cn, gaoxb@cqupt.edu.cn}

\thanks{Deyu Meng is with the National Engineering Laboratory for Algorithm and Analysis Technologiy on Big Data, Xi'an Jiaotong University,Xi'an 710049, China.	E-mail: dymeng@xjtu.edu.cn} 

\thanks{Zizhong Chen \& E. Giem is with the Department of Computer Science and Engineering, University of California Riverside, Riverside, CA, 92521. E-mail: gieme01@ucr.edu}
}

\markboth{ARXIV}
{Shell \MakeLowercase{\textit{et al.}}: Bare Demo of IEEEtran.cls for IEEE Journals}

\maketitle

\begin{abstract}
	
	This paper presents a strong data mining method based on the rough set, which can realize feature selection, classification and knowledge representation at the same time. Although the rough set, a popular method for feature selections, has a good interpretability, its low efficiency and low accuracy prevent it from immediately being applied to real-world scenarios. To address the efficiency issue of the rough set, we discover the stability of the local redundancy (SLR) of attributes and propose the theorem to rigorously prove it. In regards to the accuracy issue,  we first show that overfitting leads to the ineffectiveness of rough sets, especially when processing noise attributes. We then propose relative importance, a robust measurement for an attribute to alleviate such overfitting issues. In this paper, we propose a novel rough set framework, which significantly improves the efficiency and accuracy upon existing rough set methods. We further develop our rough set framework by proposing a "rough concept tree" concept for knowledge representation and classification. Experimental results on public benchmark data sets show that our proposed framework achieves higher accuracies than seven popular rough set methods and other state-of-the-art feature selection methods.
\end{abstract}

\begin{IEEEkeywords}
    rough set, attribute reduction, feature selection, acceleration framework
\end{IEEEkeywords}

\IEEEpeerreviewmaketitle

\section{Introduction\label{sec:introduction}}

\IEEEPARstart{F}{eature} selection, also called as attributed selection, is a foundational process in data preprocessing. Not only reducing the difficulty of learning, it also alleviates the disaster of dimensionality. Due to these advantages, past decades have witnessed rapid developments in this promising direction. To be more specific, there are three major research directions regarding feature selection: global optimal search, random search and heuristic search according to the formation process of the search strategy subset. The branch and bound method, which searches the optimal subset when the feature number in the optimal feature subset is determined in advance, is the only global search method capable of obtaining the optimum \cite{Dash1997}. In contrast, the random search strategy assigns a certain weight to each feature during the operation of the algorithm.For example, the relief series algorithm is a typical random search method that selects features according to the weight. This simple-yet-effective algorithm, although possesses broad applications, is unable to remove the redundancy and only be applicable to two classification types \cite{Pawlak1982}. On the other hand, the floating generalized backward selection method in heuristic search is a feature selection search strategy that is more conducive to practical applications. It not only considers the statistical characteristics between features, but also ensures the speed and stability of the algorithm operation \cite{Wei2010}. The sound efficiency of the heuristic search strategy comes at a cost of sacrificing the accuracy regarding the global optimum. According to whether a selection method is independent of the learning algorithm, it can be divided into three categories --- filter method, wrapper method and embedded method. The main idea of the filter method is to score the features, and then sort the features in a descending order according to their scores. The wrapper method recognizes feature selection as a feature subset optimization process, and uses classifiers to evaluate the feature subset. Since each feature subset needs to train a classifier, most wrapper methods are inefficient. Besides, limited by the classifier used by this method, the feature subset obtained by wrapper methods tends to have lower versatility. Therefore, prior works of wrapper method mainly focuses on the optimization process, motivating the embedded method, where the feature selection is embedded into the training process.

Feature selection based on the rough set theory belongs to the filter method. Rough set theory, an effective mathematical tool to process uncertain, inconsistent and incomplete data, was first proposed by a Polish scientist, Pawlak, in 1982 \cite{Pawlak1982}. It has been widely applied in machine learning, data mining and decision support systems \cite{Wei2010,Ma2014,Qian2015,Li2017,Qian2018,Vidhya2018,Hu2019}. Since the forward heuristic attribute selection algorithm based on the rough set theory can effectively reduce the time complexity, past decades witnessed rapid developments in its related fields \cite{Yao1990,Ziarko1993,Parthalain2010,Chen2012,Liang2014,Chen2014,Maji2014}. To elaborate, Miao and Hu redefined the evaluation criteria of attributes in a decision system from the perspective of mutual information using the attribute importance as a heuristic search strategy to reduce the search space during the attribute reduction \cite{miao1999heuristic}. Wang, Yu, and Yang designed a heuristic knowledge reduction algorithm for decision tables using conditional information entropy as the heuristic. They performed a careful analysis for informative and algebraic definitions in the rough set theory \cite{wang2002decision}. Liang, Chin, Dang, and Yam introduced a new definition of information entropy and the concept of complementary information entropy \cite{liang2002new}. They validated that their metric can be categorized as a fuzzy entropy, which can be used to measure the fuzzy degree and rough classification of rough sets. They also proposed a heuristic algorithm based on complementary information entropy. It is worth noting that all of these methods, which are designed for specific scenarios, cannot be used to accelerate rough set as a general method. In contrast, Qian and Liang \cite{Qian2010AI} proposed the only general method to accelerate the rough set by introducing the positive region acceleration and pre-calculation for core attributes.

In fact, the efficiency of rough set is yet to be efficient and accurate enough. Besides, as far as we know, there are few studies being capable of improving the accuracy of rough sets for a wide range of applications. These shortcomings state the major obstacles of immediate real-world deployments for an algorithm based on the rough sets. In order to address such issues, we present the following contributions in this paper:


(1) We propose a rigorously proved theorem to elaborate the stability of the attribute local redundancy in a rough set, which can be used to considerably accelerate most of the existing rough set algorithms.


(2) We discover that overfitting, especially when in a type of noise attribute, accounts for the ineffectiveness of rough sets. Our brand-new measurement increases the robustness of the rough set to the type of noise attribute with an improved generalizability while significantly enhancing its accuracy.


(3) We propose a novel general framework, RSLRS, which is not only able to considerably accelerate most of the rough set methods, but enhance their accuracies as well.


(4) We present the concept of "rough concept tree" for knowledge representations and classifications, enabling algorithms based on the rough set to effectively implement feature selection, data classification and knowledge representation simultaneously, rather than replying on any other classifiers.

The main contents of subsequent chapters are organized as follows: Chapter \Rmnum{2} lists the related work regarding the basic models of rough sets; Chapter \Rmnum{3} details the newly proposed accelerated rough set model based on the attribute local redundancy; Chapter \Rmnum{4} discusses the model of "relative importance" which can improve accuracy of feature selection of rough sets; in chapter \Rmnum{5}, an efficient and accurate rough set framework is proposed based on the proposed theorems and model; in Chapter \Rmnum{6}, we develop the rough set so that it can be effectively used for knowledge representation and data classification; the experimental results and analysis are presented in Chapter \Rmnum{7}; We conclude our paper in Chapter \Rmnum{8}.

\section{Related Work\label{sec:relatedwork}}
\subsection{Filter Methods}
The essence of filter methods is to use some indicators in mathematical statistics to score features, such as Pearson correlation coefficient, Gini coefficient, Kullback–Leibler divergence, Fisher score, Similarity measure, and etc. Since the filter method only uses the dataset rather than relying on a specific classifier, it has strong versatility and is easy to expand. Compared with wrapper methods and embedded methods, filter methods have lower algorithm complexity. However, at the same time, the classification accuracy of filter methods is usually lower than the other two methods. In addition, because filter methods only score a single feature rather than the whole feature subset, the feature subset obtained by filter methods usually has a high redundancy.


G. Qu et al\cite{Gu2012} proposed a generalized Fisher score feature selection method, which aims at finding a feature subset to maximize the lower bound of Fisher score. This method transforms feature selection into a quadratically constrained linear programming and uses the cutting plane algorithm to solve this problem. G. Roffo et al proposed a feature selection method based on graph, which ranks the most important features by recognizing them as arbitrary clue sets \cite{Roffo2017}. This method maps a feature selection problem to an affinity graph by taking features as nodes, and then evaluates the importance of nodes by the eigenvector centrality. G. Roffo et al. also proposed a filtering feature selection method, Inf-FS \cite{Roffo2020}, which takes features as nodes of a graph and views feature subset as a path in the graph. The power series property of matrix is used to evaluate the path, and the computational complexity is reduced by adding the paths until the length reaches infinity.

\subsection{Wrapper Methods}

Wrapper methods evaluate features using the performance of learning algorithms, so the classification accuracy of wrapper methods is often higher than that of filter methods. However,limited by the classifier used by this method, the feature subset obtained by wrapper methods tends to have a lower versatility. For each feature subset, wrapper methods need to train a classifier. Therefore, this method usually has a high computational complexity of, which depends on the search strategy of the feature subset. However, wrapper methods evaluate the entire feature subset, rather than a single feature, and consider the dependency between features, so the redundancy of the result feature subset is often lower than that of filter methods.


Support vector machine (SVM) is a popular learning algorithm of the wrapper method. I. Guyon et al proposed a feature selection method of support vector machine based on recursive feature elimination \cite{Guyon2002}. The idea of this method is to construct the ranking coefficient of features according to the weight vector generated by SVM during training. In each iteration, the feature with the smallest ranking coefficient is removed. At the end of this algorithm, a descending sorting of all the features is obtained. J. Guo et al proposed a feature selection method based on distance-based clustering, which uses a triplet-based ordinal locality preserving loss function to capture the local structures of the original data. Meanwhile, this method defines an alternating optimization algorithm based on half-quadratic minimization to speed up the optimization process of this algorithm \cite{Guo2017}. In order to overcome the issue that only single feature is considered in the filtering method, a joint learning framework for feature selection and clustering, namely Dependence Guided Unsupervised Feature Selection (DGUFS), was proposed \cite{Guo2018}. DGUFS is a projection-free feature selection model based on $L_{2,0}$-norm equality constraints and defines two dependence guided terms to enhance the dependence among original data, cluster labels and selected features.

\subsection{Embedded Methods}

Embedded methods embed feature selection as a component into the learning algorithm. The feature subset can be obtained when the training process of the learning algorithm is completed. The embedded method is similar to the filter method. However, it determines the score of features via model training, which is different from the filter method. The main idea of this method is to select features important to the model training when to determine the model. Meanwhile, an embedded method is compromise to a filter method and a wrapper method. Compared with the filter method, the embedded method can achieve a higher classification accuracy. When comparing to wrapper methods, embedded methods have a lower algorithm complexity and are more difficult to suffer from the overfitting issue.


P. Bradley et al proposed an embedded feature selection method based on concave minimization and support vector machine \cite{Bradley1998}, which finds a separation plane to distinguish two point sets in the n-dimensional feature space which uses as few features as possible. This method minimizes the weighted sum of the distance between the wrong classification points and the boundary plane while maximizing the distance between the two boundary planes of the separation plane. Embedded methods are often based on regression learning algorithms. F. Nie et al \cite{Nie2010} proposed an efficient and robust feature selection method, which uses a loss function based on $L_{2,1}$-norms to remove the outliers. This method adopts joint $L_{2,1}$-norms minimization on loss function and regularization and proposes an effective algorithm to solve a series of joint $L_{2,1}$-norms minimization problems. Y. Yang et al proposed a feature selection method named Unsupervised Discriminative Feature Selection (UDFS) \cite{Yang2011}. UDFS optimizes a $L_{2,1}$-norm regularized minimization loss function, which utilizes discriminative information and local structure of data distribution.

\subsection{Rough Set Theory}


The rough set theory mainly focuses on the information granulation and approximation. An equivalence relation granulates the sample space into disjoint equivalence classes, or sample information granules. It uses upper and lower approximation bounds to characterize the uncertainty of the information granules in order to approximate the arbitrary knowledge in the sample space. Classic rough set theory is defined on the basis of a strict equivalence relation and an indiscernibility relation. In this section, in order to lay a foundation for our theorem and proof, we review some of the basic concepts of they rough set theory, which has been presented in our previous work \cite{xia2020gbnrs}. Here we first introduce information systems and indistinguishable relations.

\textbf{\emph{Definition 1.}} \cite{xia2020gbnrs} An \textit{information system} is a quarternion $\left \langle U,A,V,f\right \rangle$, where:
$ U=\left \{ x_{1},x_{2},...,x_{n} \right \}$ denotes a non-empty finite set of objects. $U$ is called the universe;
$A=\left \{ a_{1},a_{2},...,a_{m} \right \}$ denotes a non-empty finite set of attributes;
$V=\bigcup_{a\in A}V_{a}$ denotes the set of all attribute values, where $V_{a}$ denotes the value range of attribute $a$;
$f=U\times A\rightarrow V$ denotes a mapping function. $\forall x_{i}\in U,a\in A $, we have $f\left ( x_{i},a \right )\in V_{a}$.
If the set of attributes $A$ in the information system above satisfies: $A=C\cup D,C\cap D=\O ,D\neq \O $, where $C$ is the conditional attribute set and $D$ is the decision attribute set, then the information system is called the \textit{decision system} $\left \langle U,C,D \right \rangle$.

\textbf{\emph{Definition 2.}} \cite{dai2018maximal-discernibility-pair-based} Let $\left \langle U,A,V,f\right \rangle$ be an information system. $ \forall x,y\in U $ and $ B\subseteq A $, the \textit{indistinguishable relationship} $ IND(B) $ of the attribute subset $ B $ is defined as
\begin{equation}
IND(B)\!=\!\{(x, y) \in U \times U | f(x, a)=f(y, a), \forall a \in B\}.
\end{equation}
In fact, $ (x, y) \in IND(B) $ shows that the values of samples $ x $ and $ y $ are completely consistent under the attribute subset $ B $; that is, under the description of the attribute subset $ B $, the samples $ x $ and $ y $ are indistinguishable.

$ IND(B) $ is symmetric, reflexive, and transitive; that is, $ \forall B\subseteq A $, $ IND(B) $ is an equivalence relation on $ U $ (abbreviated $ R_{B} $). $ IND(B) $ creates a partition of $ U $, denoted $ U/IND(B) $ and abbreviated $ U/B $. The characteristics of $ U/B $ are as follows: Suppose $ U/B=\{ X_{1},X_{2},...,X_{k} \} $, if $ X_{i}, X_{j}\subseteq U $, $ X_{i}\cap X_{j} = \O $($ i\neq j $), $ \bigcup_{i=1}^{k}X_{i}=U $, then $ U $ is divided into $ k $ parts by $ IND(B) $ and they form a partition of $ U $. An element $ [x]_{B}=\{y \in U |(x, y) \in IND(B)\} $ in $ U/B $ is called an equivalence class. This leads us to our next set of definitions, approximations based on the equivalence relation $R_B$.

\textbf{\emph{Definition 3.}} \cite{dai2018maximal-discernibility-pair-based} Let $\left \langle U,A,V,f\right \rangle $ be an information system. $ \forall B \subseteq A $ there is a corresponding equivalent relationship $ R_{B} $ on $ U $. $ \forall X \subseteq U $, the \textit{upper and lower approximations of} $ X $ with respect to $ B $ are defined as follows:
\begin{equation}
\overline{R_{B}}X=\cup\left\{[x]_{B} \in U / B |[x]_{B} \cap X \neq \emptyset\right\},
\end{equation}
\begin{equation}
\underline{R_{B}}X=\cup\left\{[x]_{B} \in U / B |[x]_{B} \subseteq X\right\}.
\end{equation}
The lower approximation $ \underline{R_{B}}X $ represents the set of samples in $ U $ that are determined to belong to $ X $ according to the equivalence relation $ R_{B} $. It essentially reflects the ability of the equivalence relation $ R_{B} $ to approximately describe the knowledge contained in $ X $ by a partition of the knowledge of the universe $ U $. It is also commonly called as the \textit{$ B $ positive region of $ X $ in $ U $}, abbreviated as $ POS_{B}(X) $ in the following contents of the paper.

\textbf{\emph{Definition 4.}} \cite{dai2018maximal-discernibility-pair-based} Let $ \left \langle U,C,D \right \rangle $ be a decision system. We notate the partition of the universe $U$ by the decision attribute set $D$ into $L$ equivalence classes by $ U/D=\left \{ X_{1},X_{2},...,X_{L} \right \}$. $ \forall B\subseteq C$, there is a corresponding equivalent relationship $ R_{B} $ on $ U $. The \textit{upper approximation and the lower approximation of $D$ with respect to $B$} are respectively defined as
\begin{equation}
\overline{R_{B}}D=\bigcup_{i=1}^{L}\overline{R_{B}}X_{i},
\end{equation}
\begin{equation}
\underline{R_{B}}D= \bigcup_{i=1}^{L}\underline{R_{B}}X_{i}.
\end{equation}

\textbf{\emph{Definition 5.}} \cite{dai2018maximal-discernibility-pair-based} Let $ \left \langle U,C,D \right \rangle$ be a decision system. $\forall B\subseteq C$, the \textit{positive region} and \textit{boundary region} of $D$ with respect to $B$ are respectively defined as:
\begin{equation}
POS_{B}\left ( D \right )=\underline{R_{B}}D,
\end{equation}
\begin{equation}
BN_{B}\left ( D \right )=\overline{R_{B}}D-\underline{R_{B}}D.
\end{equation}

The size of the positive region reflects the separability of the classification problem in a given attribute space. The larger the positive region, the more detailed the classification problem can be described with using this attribute set. We find it useful to describe it mathematically: the \textit{dependence} of $D$ on $B$ is defined as
\begin{equation}
\gamma _{B}\left ( D \right )=\frac{\left | POS_{B}\left ( D \right ) \right |}{\left | U \right |},
\end{equation}
where $\left | \cdot  \right |$ is the cardinality of the set and $0\leq \gamma _{B}\left ( D \right )\leq 1$. Obviously, the larger the positive region, the stronger the dependence of $D$ on $B$.

The dependency function defines the contribution of conditional attributes to a classification, so it can be utilized as an evaluation index for the importance of the attribute set.

\textbf{\emph{Definition 6.}} \cite{Qing2008Numerical} Given a decision system $ \left \langle U,C,D \right \rangle$, for all $B\subseteq C$ and $a\in C-B $, the \textit{importance} of $a$ relative to $B$ is defined as
\begin{equation}
SIG\left ( a,B,D \right )=\gamma _{B\cup a}\left ( D \right )- \gamma _{B}\left ( D \right ).
\end{equation}


The most widely used rough set is forward heuristic. It is not only efficient but can generate different knowledge rules as well. Therefore, the forward heuristic rough set is set as the default method in this paper. To be more specific, it uses the measurement $SIG$ in (9) to select attributes in a forward way. The selection result $C'$ is initialized with $\O$. For each attribute $a$ in the attribute $C-C'$, the one with the largest value of $SIG(a,C', D)$ and is greater than 0 will be selected into $C'$. This process is repeated until there is no more attribute greater than 0 in $SIG(a,C', D)$.

\section{Speed up Rough Set using the stability of local redundancy of an attribute\label{sec:novel}}

\subsection{Motivation\label{subsec:motivation1}}
In \cite{Qian2010AI} and \cite{RSC2000}, three most important theorems for generally speeding up the rough set that can be described as follows:

\textbf{\emph{Theorem 1}} {\it Given a decision system $ \left \langle U,C,D \right \rangle$, for $ B\subseteq C $, we have
	\begin{equation}
	POS_B(D)\subseteq POS_C(D).
	\end{equation}
}  


That is, the positive region of a attribute set must belong to the positive region of its parent attribute set. According to Theorem 1, we deduce that for any given parent attribute sets, the positive region of a child attribute set is not needed in order to calculate the positive region of the parent attribute set. Therefore, a general rough set algorithm can be accelerated.

\textbf{\emph{Theorem 2}} Let $ \left \langle U,C,D \right \rangle$ be a decision system. Let $ R'\subseteq R\subseteq C $  and $ a\in R$ be a given attribute, and let $R-R'$ be a relative reduction of $R$. If $POS_{R-a}(D) \neq POS_{R}(D) $, i.e., that $a$ is a single non-redundant attribute, $a$ is not a redundant attribute in $R-R'$. In other words, a single non-redundant attribute is a core attribute.

Theorem 2 indicates that is that the non-redundant attributes relative to given attribute set is also the non-redundant attribute relative to its child attribute set. Therefore, the non-redundant attributes in the all attributes are not needed to be considered at any time. The calculation cost of attribute combination can be reduced accordingly. We find that, from the perspective of the stability, Theorem 1 locates the stability of positive region while Theorem 2 finds the stability of non-redundancy attribute. That is, a redundant attribute relative to the child attribute set under the conditions in Theorem 1 is also the redundant attribute relative to its parent attribute set. In other words, the redundancy of the attribute is stable. Therefore, this type of redundant attributes will not be considered in the later feature selection process of a rough set, such that the rough set can be accelerated. 

\textbf{\emph{Theorem 3}} \label{theorem:SRA} \cite{RSC2000} {\bf{\{The stability of redundancy (SR) attribute\}.}} Given a decision system $ \left \langle U,C,D \right \rangle$, let $ R\subseteq C $ and $ a\in C $ is a given attribute. If $ U/R = U/(R + a) $, $a$ is a redundant attribute relative to $ R$. 


Theorem 3 describes a finding on redundancy attributes for further speeding up the rough set. However, the condition to decide whether a attribute is redundant or not is rigorous. Therefore, in some cases, one may not be able to observe a significant acceleration. In the following theorem, we find that, even for most of those non-redunant attributes, there exists the local redundancy so some of the calculations for the equivalence class can be avoided. The theorem related to the local redundancy is detailed in the next section.

\subsection{The Stability of Local Redundancy of an Attribute for accelerating Rough Set \label{subsec:novel1}}


Inspired by Theorem 3, we describe the definitions of the active region and the non-active region.

\textbf{\emph{Definition 7.}} {\bf{\{Active Region and Non-Active Region\}.}} Let $ \left \langle U,C,D \right \rangle $ be a decision system. Let $ R\subseteq C $ and $ a\in C $ (but $ a\notin R $) be a given attribute. The equivalence class that $ U $ is divided into under the attribute set $ R $ is $ U/R = \left \{X_{1},X_{2},...,X_{i},X_{i+1},...,X_{l} \right \} $ and the equivalence class that $ U $ is divided into under the attribute $ a $ is $ U/a = $\{$X_{1}^{'},X_{2}^{'},...,X_{k}^{'},$ $X_{k+1}^{'},...,X_{s}^{'}$\}. For $\forall$  $j\in{1,2...,i}$, if $ X_{j} \subseteq X_{t}^{'} $($ t=1,2,...,s $) exists, then we define set $ NACT_{a}(R)=X_{1}\cup X_{2}\cup ...\cup X_{i} $ as the non-active region of attribute $ a $ relative to $R$ and set $ ACT_{a}(R)=X_{i+1}\cup ...\cup X_{l} $ as the active region of the attribute $ a $ relative to $R$.

Based on Definition 7 and Theorem 3, a better acceleration method is presented in Theorem 4.

\textbf{\emph{Theorem 4}}  {\it {\bf{\{The stability of local redundancy (SLR) of attributes\}.}} Given a decision system $ \left \langle U,C,D \right \rangle$, let $ R\subseteq C $ and $ a\in C $ (but $ a\notin R $) be a given attribute. The equivalence class that $ U $ is divided into under the attribute set $ R $ is $ U/R = \left \{X_{1},X_{2},...,X_{i},X_{i+1},...,X_{l} \right \} $ and the equivalence class that $ U $ is divided into under the attribute $ a $ is $ U/a = $\{$X_{1}^{'},X_{2}^{'},...,X_{k}^{'},$ $X_{k+1}^{'},...,X_{s}^{'}$ \}. If $X_{j} \subseteq X_{t}^{'} ( j = 1,2,...,i ,  t=1,2,...,s )$, the active region of $ a $ relative to $R$ on $U$ is represented as $ ACT_{a}(R) = X_{i+1}\cup ...\cup X_{l} $, and the non-active region of $ a $ relative to $R$ as $ NACT_{a}(R) = X_{1}\cup X_{2}\cup ...\cup X_{i}$. We only need to pay attention to $ACT_{a}(R)$ to determine whether $ a $ is a non-redundant attribute relative to the attribute set $ R $.} 

\noindent
{\bf Proof}. let $ U/R = \left \{X_{1},X_{2},...,X_{i},X_{i+1},...,X_{l} \right \} $ , $ NACT_{a}(R) = X_{1}\cup X_{2}\cup ...\cup X_{i}$, $ ACT_{a}(R) = X_{i+1}\cup ...\cup X_{l} $.

Since $ NACT_{a}(R) $ is the non-active region of $ a $ relative to $R$, we have
\begin{equation}
X_{j} \subseteq X_{t}^{'} ( j = 1,2,...,i ,  t=1,2,...,s ),
\end{equation}
\begin{equation}
X_{j} \cap X_{t}^{'} = X_{j},
\end{equation}
\begin{equation}
\begin{aligned}
U/(R + a) = \{X_{1}\cap X_{1}^{'}, X_{1}\cap X_{2}^{'},..., X_{2}\cap X_{1}^{'}, \\ 
X_{2}\cap X_{2}^{'},..., X_{i}\cap X_{1}^{'},..., X_{l}\cap X_{s}^{'} \} \\
= \left \{X_{1},X_{2},...,X_{i},X_{i+1}\cap X_{1}^{'},X_{i+1}\cap X_{2}^{'},..., X_{l}\cap X_{s}^{'} \right \}\\
= \left \{NACT_{a}(R),X_{i+1}\cap X_{1}^{'},X_{i+1}\cap X_{2}^{'},..., X_{l}\cap X_{s}^{'} \right \}.
\end{aligned}
\end{equation}

Focusing only on $ ACT_{a}(R) $, i.e., that equivalence classes are divided on $ ACT_{a}(R) $ and not on $NACT_{a}(R)$, we have, 
\begin{equation}
\begin{aligned}
U/(R+a) = NACT_{a}(R) + ACT_{a}(R)/(R+a) \\
= NACT_{a}(R) + \{X_{i+1}\cap X_{1}^{'},X_{i+1}\cap X_{2}^{'},..., X_{l}\cap X_{s}^{'}\}\\
= \{NACT_{a}(R), X_{i+1}\cap X_{1}^{'},X_{i+1}\cap X_{2}^{'},..., X_{l}\cap X_{s}^{'} \},
\end{aligned}
\end{equation}
which is the same as (26).

$ \Rightarrow $ Focusing only on $ ACT_{a}(R) $ can determine the redundancy of $ a $ relative to the attribute set $ R $.\\


Theorem 4 shows that, regardless of whether an attribute is redundant or not, its non-active region does not need to participate into further calculations, which results in a significant improvement for the performance. Here we use an example to demonstrate the theorem. Supposing that $ U=\left \{ x_{1},x_{2},...,x_{8} \right \}$ is a universe, $C=\left \{ a_{1},a_{2},...,a_{6} \right \}$ is a set of conditional attributes containing six attributes and $ D=\left \{ d \right \} $ is a set of decision attributes. there is a decision system $\left \langle U,C,D \right \rangle$, as shown in Table \ref{table_1}.

\begin{table}[htp]
	\renewcommand{\arraystretch}{1.3}
	\caption{Decision System $\left \langle U,C,D \right \rangle$}
	\centering	
	\label{table_1}
	\begin{tabular}{cccccccc}
		\hline
		& $ a_1 $ & $ a_2 $ & $ a_3 $ & $ a_4 $ & $ a_5 $ & $ a_6 $ & $ d $ \\
		\hline
		$ x_1 $ & 1  & 0  & 1  & 1  & 0  & 1  & 1 \\
		$ x_2 $ & 0  & 0  & 1  & 1  & 0  & 1  & 0 \\
		$ x_3 $ & 1  & 1  & 0  & 1  & 0  & 1  & 0 \\
		$ x_4 $ & 1  & 1  & 0  & 1  & 0  & 1  & 1 \\
		$ x_5 $ & 0  & 0  & 0  & 0  & 1  & 0  & 0 \\
		$ x_6 $ & 1  & 0  & 1  & 1  & 0  & 0  & 0 \\
		$ x_7 $ & 1  & 0  & 0  & 0  & 1  & 1  & 1 \\
		$ x_8 $ & 1  & 1  & 1  & 0  & 0  & 0  & 1  \\
		\hline
	\end{tabular}
\end{table}

From Table \ref{table_1}, we can calculate the following results. \\
$ U/D=\left \{ \left \{ x_{1},x_{4},x_{7},x_{8} \right \},\left \{ x_{2},x_{3},x_{5},x_{6} \right \} \right \} $, \\
$ U/C\!=\!\left \{ \left \{ x_{1} \right \},\left \{ x_{2} \right \}, \left \{ x_{3},x_{4} \right \}, \left \{ x_{5} \right \}, \left \{ x_{6} \right \}, \left \{ x_{7} \right \}, \left \{ x_{8} \right \} \right \} $,\\
$ POS_{C}(D)=\left \{ x_{1},x_{2},x_{5},x_{6},x_{7},x_{8} \right \} $.\\
$ U\!/\!(C\!-\!\left \{ a_{1} \right \})\! =\! \left \{\! \left \{ x_{1}\!,\! x_{2} \right \}\!,\! \left \{ x_{3}\!,\!x_{4} \right \}\!,\! \left \{ x_{5} \right \}\!,\! \left \{ x_{6} \right \}\!,\! \left \{ x_{7} \right \}\!,\! \left \{ x_{8} \right \}\! \right \} $,\\
$ POS_{C-\left \{ a_{1} \right \}}(D)=\left \{ x_{5},x_{6},x_{7},x_{8} \right \} \neq POS_{C}(D) $.\\
$ U\!/\!(C\!-\!\left \{ a_{2} \right \})\! =\! \left \{\! \left \{ x_{1} \right \}\!,\! \left \{ x_{2} \right \}\!,\! \left \{ x_{3}\!,\!x_{4} \right \}\!,\! \left \{ x_{5} \right \}\!,\! \left \{ x_{6} \right \}\!,\! \left \{ x_{7} \right \}\!,\! \left \{ x_{8} \right \}\! \right \} $,\\
$ POS_{C-\left \{ a_{2} \right \}}(D)=\left \{ x_{1},x_{2},x_{5},x_{6},x_{7},x_{8} \right \}=POS_{C}(D) $.\\
$ U\!/\!(C\!-\!\left \{ a_{3} \right \})\! =\! \left \{\! \left \{ x_{1} \right \}\!,\! \left \{ x_{2} \right \}\!,\! \left \{ x_{3}\!,\!x_{4} \right \}\!,\! \left \{ x_{5} \right \}\!,\! \left \{ x_{6} \right \}\!,\! \left \{ x_{7} \right \}\!,\! \left \{ x_{8} \right \}\! \right \} $,\\
$ POS_{C-\left \{ a_{3} \right \}}(D)=\left \{ x_{1},x_{2},x_{5},x_{6},x_{7},x_{8} \right \}=POS_{C}(D) $.\\
$ U\!/\!(C\!-\!\left \{ a_{4} \right \})\! =\! \left \{\! \left \{ x_{1} \right \}\!,\! \left \{ x_{2} \right \}\!,\! \left \{ x_{3}\!,\!x_{4} \right \}\!,\! \left \{ x_{5} \right \}\!,\! \left \{ x_{6} \right \}\!,\! \left \{ x_{7} \right \}\!,\! \left \{ x_{8} \right \}\! \right \} $,\\
$ POS_{C-\left \{ a_{4} \right \}}(D)=\left \{ x_{1},x_{2},x_{5},x_{6},x_{7},x_{8} \right \}=POS_{C}(D) $.\\
$ U \! /\! (C \! -\! \left \{ a_{5} \right \})\! =\! \left \{\! \left \{ x_{1} \right \}\!,\! \left \{ x_{2} \right \}\!,\! \left \{ x_{3}\!,\! x_{4} \right \}\!,\! \left \{ x_{5} \right \}\!,\! \left \{ x_{6} \right \}\!,\! \left \{ x_{7} \right \}\!,\! \left \{ x_{8} \right \}\! \right \} $,\\
$ POS_{C-\left \{ a_{5} \right \}}(D)=\left \{ x_{1},x_{2},x_{5},x_{6},x_{7},x_{8} \right \}=POS_{C}(D) $.\\
$ U\!/\!(C\!-\!\left \{ a_{6} \right \})\! =\! \left \{\! \left \{ x_{1}\!,\!x_{6} \right \}\!,\! \left \{ x_{2} \right \}\!,\! \left \{ x_{3}\!,\!x_{4} \right \}\!,\! \left \{ x_{5} \right \}\!,\! \left \{ x_{7} \right \}\!,\! \left \{ x_{8} \right \} \! \right \} $,\\
$ POS_{C-\left \{ a_{6} \right \}}(D)=\left \{ x_{2},x_{5},x_{7},x_{8} \right \} \neq POS_{C}(D) $.

According to the above results, the core attribute set can be obtained as $ \left \{ a_{1},a_{6} \right \} $. \\
$ U/\left \{ a_{1},a_{6} \right \}=\left \{ \left \{ x_{1},x_{3},x_{4},x_{7} \right \},\left \{ x_{2} \right \}, \left \{ x_{5} \right \}, \left \{ x_{6},x_{8} \right \} \right \} $,\\
$ POS_{\left \{ a_{1},a_{6} \right \}}(D)=\left \{ x_{2},x_{5} \right \} $.\\
$ U/a_{2}=\left \{ \left \{ x_{1},x_{2},x_{5},x_{6},x_{7} \right \},\left \{ x_{3},x_{4},x_{8} \right \} \right \} $,\\
$ \left \{ x_{2} \right \}\subseteq \left \{ x_{1},x_{2},x_{5},x_{6},x_{7} \right \} $, $ \left \{ x_{5} \right \}\subseteq \left \{ x_{1},x_{2},x_{5},x_{6},x_{7} \right \} $.

According to Definition 7, as $NACT_{a_2}(\{a_{1}, a_{6}\})= \left \{ x_{2},x_{5} \right \} $ , Theorem 1 can be used for acceleration and Theorem 4 is not needed. Thus, we use $a_3$ as an example to show the effectiveness of Theorem 4. Also based on Definition 7,  $NACT_{a_3}(\{a_{1}, a_{6}\})= \left \{ x_{2},x_{5},x_{6},x_{8} \right \} $, and the active region of attribute $ a_3 $ is $ ACT_{a_3}(\{a_{1}, a_{6}\})=\left \{ x_{1},x_{3},x_{4},x_{7} \right \} $. $\{x_{2}, x_{5}\}$ does not need to be considered in further calculations according to Theorem 1 because both of them belong to the positive region. $\{x_{6},x_{8}\}$ do not need to be considered in later calculations according to Theorem 4 because they belong to $NACT_{a_3}(\{a_{1}, a_{6}\})$. Thus, the number of calculations is decreased after applying Theorem 4.


In addition, we compute $POS_{\{a1,a6,a3\}}(D)$ according to whether $\{x_{6},x_{8}\}$ are paid attention. If $\{x_{6},x_{8}\}$ are paid attention, $POS_{\{a1,a6,a3\}}(D) = \{x_1\}$; otherwise, $POS_{\{a1,a6,a3\}}(D) = \{x_1\}$. Thus, $POS_{\{a1,a6,a3\}}(D)$ remains to be the same whether $\{x_{6},x_{8}\}$ are paid attention or not. So, this case demonstrates the effectiveness of Theorem 4.

\section{A robust measurement of evaluating the importance of an attribute for improving accuracy of rough set}
\subsection{Motivation}


In addition to the unfavorable performance, another reason preventing the rough set from being widely used is its less competitive classification accuracy. In this section, we show that the rough set is not robust to one kind of noise attribute, which widely exists in real data sets to different degree. This phenomenon can be described as follows.

\begin{table}[htp]
	\renewcommand{\arraystretch}{1.3}
	\caption{Decision System $\left \langle U,C,D \right \rangle$}
	\centering 
	\label{tb:relativePR}
	\begin{tabular}{cccccccc}
		\hline
		& $ C_1 $ & $ C_2 $ & $ C_3 $ & $ C_4 $ & $ D $ \\
		\hline
		$ $ & 1  & 3  & 4  & 5  & 1 \\
		$ $ & 2  & 3  & 4  & 5  & 1 \\
		$ $ & 3  & 4  & 3  & 2  & 1 \\
		$ $ & 4  & 3  & 3  & 2  & 0 \\
		$ $ & 5  & 5  & 5  & 5  & 0 \\
		$ $ & 6  & 4  & 3  & 2  & 1 \\
		$ $ & 7  & 4  & 3  & 2  & 1 \\
		$ $ & 8  & 3  & 3  & 2  & 0 \\
		$ $ & 9  & 5  & 5  & 5  & 1 \\
		\hline
	\end{tabular}
\end{table}


Table \ref{tb:relativePR} consists of the information about a group of students. As shown in this table, $C_{1}, C_{2}, C_{3}$ and $C_{4}$ are conditional attributes corresponding to the grades of the students. $D$ is the decision attribute telling whether a student's overall performance is qualified. One can observe that $C_{1}$ containing a sequence of values is a noise attribute, which may come from the negligence of the statistician. However, in a rough set, this attribute can recognize each sample as a equivalence and a positive region sample, resulting in the overfitting issue. Therefore, in the reduction result, there only remains this attribute. Obviously, this reduction result is incorrect, and the equivalence rules of this case does not make sense --- $C_{1}$ is such an extreme case. However, regardless of the presence of noise attributes, this phenomenon will occur to various degrees and lead to overfitting in the rough set. The specific reason is shown in Fig.1.


Fig.1(a) consists of five equivalences, seven positive region points and two boundary region points. After the attribute $C_{1}$ is added, although the number of positive region points increases from seven to nine, the number of equivalence classes increases from five to nine. It can be observed that the increased equivalence classes from the existing positive region equivalence classes are meaningless for the decision. In other words, more positive region points can decrease the boundary region, narrow the decision boundary and make the boundary clearer. This explains why the existing rough set methods utilize the number of positive region points as the measurement to evaluate the importance of an attribute. However, as shown in Fig. 1, more equivalence classes mean more rules and more complex boundaries, which is harmful for decision. It may lead to the overfitting issue, which is never considered in the existing rough set methods. Considering this problem, the relative importance is proposed to evaluate the importance of an attribute in the rough set. We consider both the number of positive region points and the number of equivalence classes.

\begin{figure}[!ht] 
	\label{fg:relativePR}
	\centering
	\subfigure[]{\includegraphics[width = 0.214\textwidth]{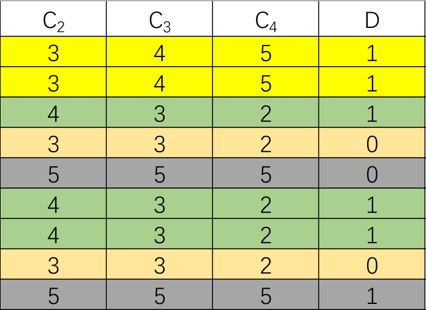}}
	\subfigure[]{\includegraphics[width = 0.267\textwidth]{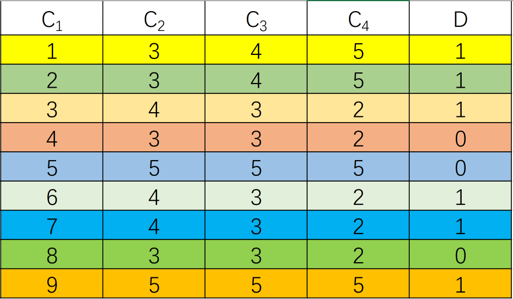}}
	\caption{The equivalence results on the data set in Table (a) the equivalence results when noise attribute $C_{1}$ is not added. (b)the equivalence results when noise attribute $C_{1}$ is added. Blocks in the same color presents one equivalence belongs to the positive region. Grey represents the boundary region. }
	\label{Fig}
\end{figure}

\subsection{Relative importance of an attribute \label{subsec:novel5}}

\textbf{\emph{Definition 8.}} Let $ \left \langle U,C,D \right \rangle$ be a decision system. $\forall B\subseteq C$ and $a\in C-B $, the \textit{relative importance} of $a$ relative to $B$ is defined as, 
\begin{equation}
\begin{split}
RSIG\left ( a,B,D \right )=\frac{\Delta \left |POS_{B}\left ( D \right )  \right |}{\Delta \left | U/B  \right |} \\
= \frac{\left |POS_{B+a}\left ( D \right )  \right |-\left |POS_{B}\left ( D \right )  \right |}{\left | U/(B+a)  \right |-\left | U/(B)  \right |},
\end{split}
\end{equation}
where $|.|$ denotes the number of the elements in a set. $\Delta$ represents the increased amount. As shown in Definition 8, both the number of positive region points and equivalence classes can be considered. To be more specific, more positive region points and fewer equivalence classes provide us with a better result. For the first attribute selection, the initial number of positive region points and equivalence classes are both equal to 0. Because the acceleration strategies in Theorems 2 may affect the denominator of the relative importance model, we do not apply them when adopting the relative importance model.

\section{An efficient and accurate rough set framework}

\begin{figure}[htb]
	\centering
	\includegraphics[width = 0.5\textwidth]{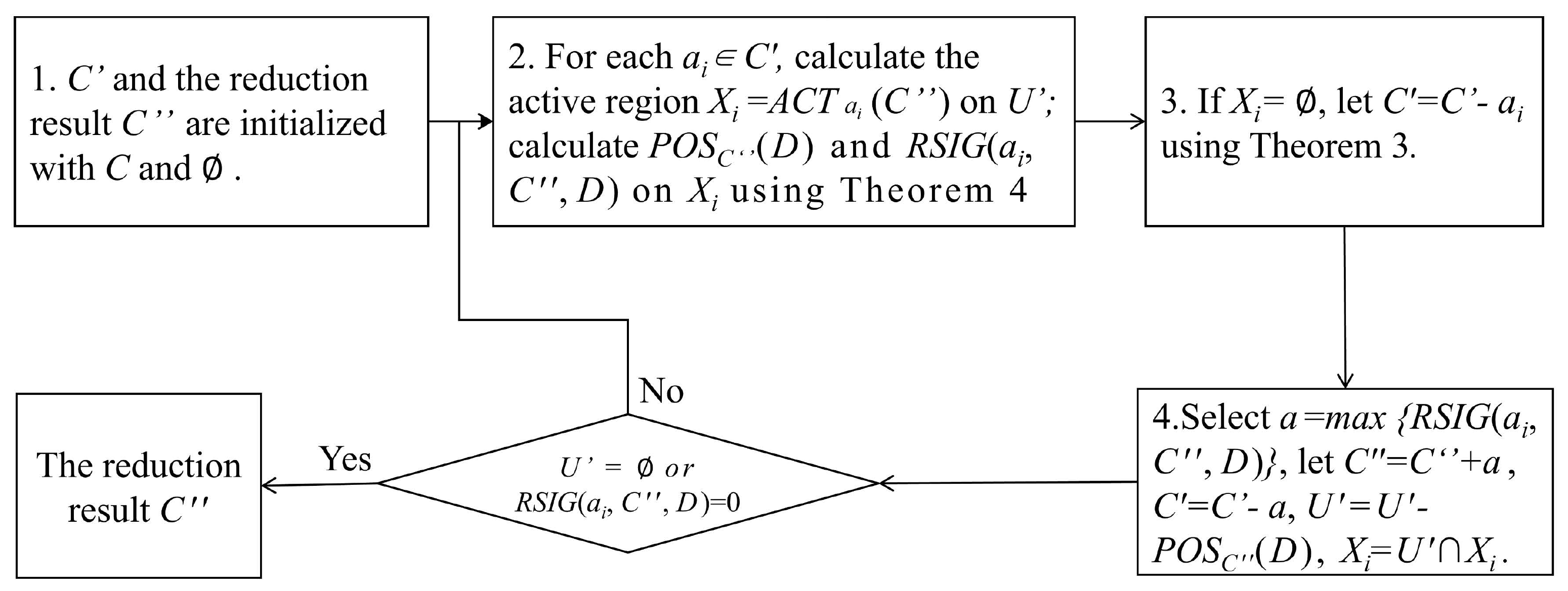}
	\caption{A flow chart of the RSLRS framework}
	\label{fig:FlowChart}
\end{figure}

Based on the aforementioned theorems regarding the stability of redundancy attributes and the relative importance model, we propose the novel RSLRS framework to accelerate rough set methods. Here we provide a definition useful to structure our framework. As shown in Fig. \ref{fig:FlowChart}, there are five main steps of our framework.

The first step is to initialize $  C^{'}=C, C^{''}=\O$, $ U^{'}=U-POS_{C^{''}}(D) $, and then comes the second step. For $ \forall a_{i}\in C^{'} $, we calculate the active region $ ACT_{c_{i}}{U^{'}}$ and  $ RSIG(c_{i}, C^{''}, D) $ on $ ACT_{c_{i}}{U^{'}}$. According to Theorem 4, in the next step, we only need to decide whether there is a change in $ ACT_{c_{i}}{U^{'}}$, so as to reduce the repeated determination of samples. 

Regarding the third step, if $ ACT_{a_{i}}{U^{'}} = \O $, we remove $ a_{i} $ from $ C^{'} $ using Theorem 3, obtaining that $ C^{'}=C^{'}-a_{i} $. According to Theorem 3, the attribute $ a_{i} $ does not help to distinguish $ U' $, thus it is a redundant attribute relative to the reduction result. By removing $ a_{i} $ from $ C' $, we reduce the number of attributes in subsequent iterations. At this step, we achieve a further acceleration on the reduction algorithm for the classical rough set attributes.

At the fourth step, we select an attribute $ a $ for $C'$, which has the largest value of $ RSIG(a_{i}, C^{''}, D) $. A larger value of $ RSIG(a_{i}, C^{''}, D) $ will lead to a stronger representation ability. We then add the selected attribute $ a $ to the non-redundant attribute set $ C^{''} $ and remove $ a $ from $ C^{'} $, such that $ C^{''}=C^{''}$ $\cup$ $ a $ and $ C^{'}=C^{'}-a $. Thereafter, we calculate the lower approximation of the updated $ C^{''} $ on $ U^{'} $ (i.e., the positive region of $ C^{''} $ on $ U^{'} $) and update the data set $ U^{'} $ using Theorem 1 such that $ U^{'}=U^{'}-POS_{C^{''}}(U^{'}) $. On the basis of updating the data set $ U^{'} $, we update $ ACT_{c_{i}}{U^{'}} $ and have $ ACT_{c_{i}}{U^{'}}=U^{'}\cap ACT_{c_{i}}{U^{'}} $. Therefore, the active region of an attribute is decreased while the later calculation is further accelerated.

After performing the above four main steps, we proceed to the next step based on whether $ U^{'} $ or $ RSIG(a_{i}, C^{''}, D) $ is an empty set or not. If $ U^{'} $ is not an empty set, we repeat step 2 as well as the following steps until either of the two conditions can be reached. Following the flowchart outlined in Fig. \ref{fig:FlowChart}, we present the algorithmic details in Algorithm \ref{alg:Alg1}, where we propose the rough set using the model of relative importance and stability of local redundancy of attributes. We call it as relative stability of local redundancy rough set (RSLRS). 

\floatname{algorithm}{Algorithm }  
\renewcommand{\algorithmicrequire}{\textbf{ Input:}}  
\renewcommand{\algorithmicensure}{\textbf{ Output:}}
\begin{algorithm}  
	\caption{Attribute reduction algorithm}
	\label{alg:Alg1} 
	\begin{algorithmic}[1] 
		\REQUIRE
		A decision system $ \left \langle U,C,D \right \rangle $;
		\ENSURE
		The reduction result $ C^{''} $;\\
		
		\STATE Initialize $  C^{'}=C, C^{''}=\O  $;
		\FOR { $ a_{i} $ $ \in $  $ C' $ }
		\STATE Calculate $ACT_{a_{i}}({C^{''}})$, and both $POS_{a_{i}+C^{''}}(D)$ and $RSIG(a_{i},C^{''},D)$ on $ACT_{a_{i}}({C^{''}})$; //Apply Theorem 4.
		\IF {$ ACT_{a_{i}}{(C^{''})}=\O $}
		\STATE $ C^{'} = C^{'} - a_{i} $; //Apply Theorem 3.
		\ENDIF
		\ENDFOR

		\STATE Select $ a = max\{(RSIG(a_{i},C^{''},D))\} $; //Select the attribute with the biggest relative importance from among the remaining attributes.

		\IF {$ RSIG(a,C^{''},D) > 0 $}
		\STATE $  C^{''}= C^{''} \cup \left \{ a \right \} $;
		\STATE $ C^{'} = C^{'} - a $;		
		\STATE $ U=U-POS_{C^{''}}(D) $;	//Remove the positive region under $a$ from $U$ using Theorem 1.
		\ELSE
		\STATE Return $ C^{''} $.
		\ENDIF
		\IF {$ U \neq \O $}
		\STATE Go to Step 2;
		\ELSE 
		\STATE Return $ C^{''} $.
		\ENDIF
	\end{algorithmic}
\end{algorithm}

\section{Knowledge represent and data classification using rough set\label{sec:classification}}

\begin{figure}[htb]
	\centering
	\includegraphics[width = 0.45\textwidth]{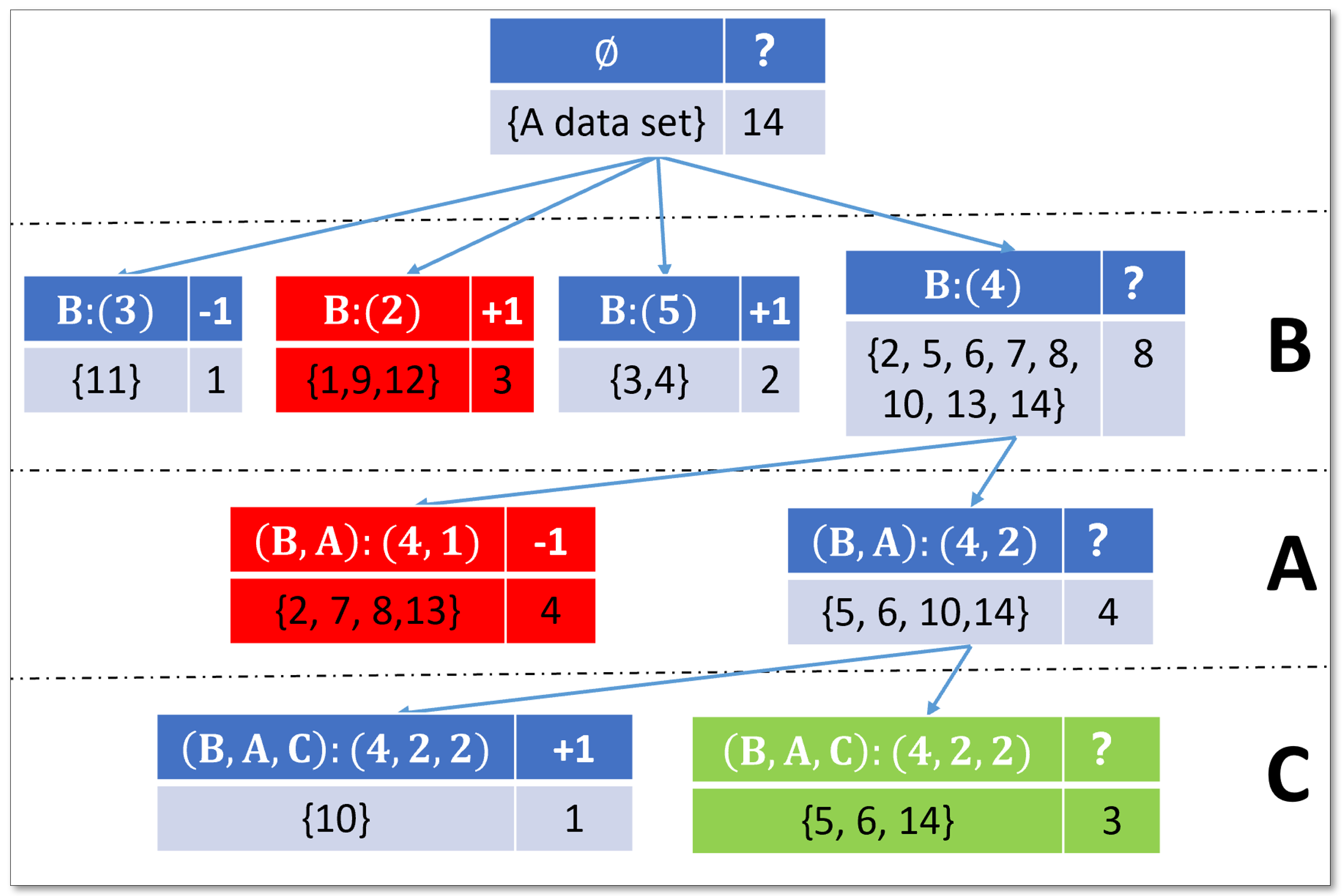}
	\caption{Knowledge representation of rough concept tree and rough set classifier}
	\label{fig:RCL}
\end{figure}

Rough sets can extract and generate knowledge rules. However, at present, there is no effective way to organize or to express the knowledge rules obtained by rough sets. To remedy this deficit, we present a knowledge representation method for rough set in this section. Take the data set of a disease as an example. The reduction result using Algorithm 1 is shown in Table 1, in which only three attributes (i.e., A, B and C) are selected. The importance of them are B, A and C respectively. The three attributes in the data set represent the physical sign level data, and the +1 and -1 in the decision attribute indicate being diseased or non-diseased. Inspired by the concept lattice \cite{Conceptlattice2010}, we propose the knowledge representation called rough concept tree(RCT) for rough sets, as shown in Fig. \ref{fig:RCL}. In the RCT, each node represents a "knowledge point" or "concept node", which consists of two parts --- a sequence with both attributes whose values are called as "intent", and its corresponding equivalence called "extent" as the second part. These two parts are corresponding to that in the first row, first column and that in the second row, first column, respectively. In each "concept node", the "extent" is corresponding to the objects in an equivalence and the "intent" is corresponding to some attributes. Their values are borrowed from concept lattice. Besides, in a concept node, the value in its second row and second column counts the number of objects in its extent, and the value in its first row and second column represents the label of the extent, i.e., the equivalence class. If this value is a number, it suggests that the equivalence class belongs to the positive region and this concept node can describe a certain knowledge. Otherwise, the value is a question mark, indicating that the equivalence class belongs to the boundary region and can not describe a certain knowledge. As for the second layer of RCT in Fig. 3 corresponding to the attribute B, the second node represents a knowledge point (or a "concept node") which consists of an extent (i.e., an equivalence class) with the 1$^{th}$, 9$^{th}$ and 12$^{th}$ points when the value of the attribute B is equal to 2, and the number of objects in the equivalence class is equal to 3. The equivalence class belongs to the positive region and its label is equal to "+1". Similarly, the second node in the third layer represents a knowledge point with an equivalence class containing the 5$^{th}$, 6$^{th}$, 10$^{th}$ and 14$^{th}$ points when the value of the attributes B and A are equal to 4 and 2, respectively. The question mark in this node represents that the equivalence class belongs to the boundary region. The RCT is ordered by the importance of attributes. The higher an attribute is, the greater the importance will be. Since the RCT is ordered, the extent corresponding to a given intent is fixed. 

\begin{table}[htp]
	\renewcommand{\arraystretch}{1.2}
	\caption{Take a disease data as an example}
	\label{tab:table_data}
	\centering
\resizebox{9cm}{0.8cm}{
	\begin{tabular}{lllllllllllllll}
				
		\hline
		Objects       &1& 2 & 3 & 4 & 5 &6 &7 &8 &9 &10 &11 &12 &13 &14 \\
		\hline
		A             &1&1&1&1&2&2&1&1&1&2&1&1&1&2 \\
		\hline
		B             &2&4&5&5&4&4&4&4&2&4&3&2&4&4 \\
		\hline
		C             &1&1&1&1&3&3&1&1&1&2&1&1&1&3 \\
		\hline
		Label         &+1&-1&+1&+1&-1&-1&-1&-1&+1&+1&-1&+1&-1&+1\\		
		\hline

	\end{tabular}
	}
\end{table}

In a CRT, as shown in Fig. \ref{fig:RCL}, all non-leaf nodes belong to boundary region, i.e., the nodes containing a question mark, indicating that the labels of the objects in a non-leaf node are not identical. Some leaf nodes belongs to the positive region, and others belongs to the boundary region, such as the green node in Fig. 3. In classification problems, as mentioned above, the leaf nodes belonging to the positive region can clearly describe the certain knowledge. The more objects in the extent and the fewer attributes in the intent in a concept node, the stronger the corresponding knowledge representation ability will be. Take Fig. 3 as an example. The two red nodes have the highest knowledge representation ability. On the one hand, regarding the two equivalence classes, the extents in the two nodes can describe the largest collection of objects, so their generalizabilities are among the best; On the other hand, the fewer attributes can lead to a higher generalizability when other conditions are close to each other. The scale of rough concept tree is much smaller than the concept lattice, which overcomes the inefficiency of the concept lattice method. In summary, we provide a rough concept tree to organize and describe the knowledge rules obtained from forward rough sets. In addition, it is worth noting that, to be convenient for display in some large-scale RCTs, the names of attributes in the first row and first column of each concept node can be neglected because the RCT is ordered as shown in Fig. \ref{fig:zooRCT}, and the attributes can thus be easily derived.


As shown in Fig. 3, RCT can be considered as a special decision tree. The rough set can be used for decision in classification problem according to the decision process, although RCT is not a rigours decision tree because its nodes contain intent and extent instead of only objects. However, in the previous work \cite{RSC2000}, since only the equivalence classes in the positive region classes describe certain knowledges, they are used for decision in classification. However, this will not utilize the uncertain information in equivalence classes belonging to boundary region though the information is also useful. Therefore, its classification accuracy is low. In this paper, learning from the general decision tree algorithm, we also use knowledge information of the boundary region for decision by introducing the voting mechanism. Using the green node belonging to boundary region as an example, its decision label equals to the majority label in its extent, i.e., -1.

Another challenge of the classification problem using rough sets is that how a classification task is completed using the RCT when any values of a new point in a test data set do not appear in the training set. For example, we need to predict the label of a person whose data values on attributes A, B and C are equal to (4.2,2.3,3) respectively using the RCT in Fig. 3. However, the value 4.2 does not appear in the attribute A, and the value 2.3 does not appear in the attribute B. To address this problem, for a value that does not appear in the existing data, we can use the closest value to approximate and replace it. Given a data set $U$ containing $n$ points and $d$ dimensionality and a point $(C_{1}^i, C_{2}^i, ..., C_{d}^i) \in U$, for an unknown point $(C_{1}, C_{2}, ..., C_{d})$, its values can be transformed using the follow mathematical model:

\begin{equation}
\scriptstyle f(C_1,C_2\dots C_d)=(C_1^*,C_2^*\dots C_d^*),s.t.
\begin{cases}
\scriptstyle C_1^*=min \,\{ dis(C_1^i,C_1)\}\\
\scriptstyle C_2^*=min \,\{dis(C_2^i,C_2)\}\\
\quad \quad  \cdots\cdots  \\
\scriptstyle C_d^*=min \,\{dis(C_d^i,C_d)\}\\
\end{cases}
,i=1,2\dots n,
\end{equation}


where $(C_{1}^{i}, C_{2}^{i}, ..., C_{d}^{i}) \in U, i=1,2...,n$. Using the value 4.2 in attribute A as an example, its closest value in attribute A is equal to 4, and in attribute B is equal to 2. Therefore, (4.2, 2.3, 3) is transformed to be as (4,2,3), and its label is equal to -1 according to RCT in Fig. 3 using the voting mechanism. The rough set classifier (RSC) is a rare classifier with good interpretability. As far as we know, logistic regression (LR) is the only classifier with this characteristic. Considering that LR is widely used in financial sector and other field, we include LR for comparison with RSC in the next section. In addition, RSC can provide an interpretability different from LR using the RCT shown in Fig. 3.

\section{Experiments\label{sec:experiments}}

\begin{table*}[htp]
	\renewcommand{\arraystretch}{1.3}
	\caption{Classification accuracy of FRPOS}
	\label{tab:FSAccurcy}
	\centering
	\resizebox{\textwidth}{!}{
		\begin{tabular}{lllllllll}
			\hline
			& cfs           & fisher                 & ilfs                   & laplacian     & lasso         & mutinffs               & RSLRS                  & CRS           \\
			\hline
			anneal                  & 0.9249±0.0161 & 0.9249±0.0161          & 0.9311±0.0161          & 0.9249±0.0161 & 0.9086±0.0200 & 0.9249±0.0161          & \textbf{0.9600±0.0151} & 0.8609±0.0097 \\
			Healthy\_Older\_People2 & 0.9717±0.0113 & 0.9724±0.0112          & 0.9717±0.0113          & 0.9311±0.0277 & 0.9311±0.0277 & 0.9724±0.0112          & \textbf{0.9739±0.0116} & 0.9561±0.0145 \\
			hepatitis               & 0.8134±0.0546 & 0.8071±0.0639          & 0.8001±0.0564          & 0.8134±0.0546 & 0.8134±0.0546 & 0.8071±0.0639          & \textbf{0.8326±0.0746} & 0.7867±0.0800 \\
			htru2                   & 0.9706±0.0042 & 0.9734±0.0045          & 0.9713±0.0043          & 0.9713±0.0044 & 0.9736±0.0037 & 0.9734±0.0045          & \textbf{0.9761±0.0037} & 0.9759±0.0033 \\
			lymphography            & 0.6964±0.1158 & 0.7954±0.0938          & 0.8095±0.0885          & 0.7254±0.1245 & 0.7593±0.0902 & 0.7954±0.0938          & \textbf{0.8353±0.0935} & 0.7961±0.1324 \\
			zoo                     & 0.8281±0.0877 & 0.8700±0.0806          & 0.8955±0.0957          & 0.8775±0.0867 & 0.8193±0.0780 & 0.8700±0.0806          & \textbf{0.9214±0.0880} & 0.8833±0.0542 \\
			letter                  & 0.6908±0.0083 & 0.6250±0.0111          & 0.6827±0.0117          & 0.5462±0.0128 & 0.6540±0.0119 & 0.6250±0.0111          & \textbf{0.7148±0.0105} & 0.6539±0.0107 \\
			GINA                    & 0.8512±0.0242 & \textbf{0.8718±0.0222} & 0.8668±0.0199          & 0.8281±0.0202 & 0.8322±0.0189 & 0.8674±0.0166          & 0.8135±0.0243          & 0.7380±0.0274 \\
			ionosphere.csv          & 0.8415±0.0655 & 0.8386±0.0605          & \textbf{0.8446±0.0657} & 0.8445±0.0722 & 0.8384±0.0658 & 0.8386±0.0605          & 0.7813±0.0623          & 0.7813±0.0623 \\
			sensorReadings.csv      & 0.5565±0.0545 & 0.6241±0.0321          & \textbf{0.6454±0.0385} & 0.6146±0.0240 & 0.6252±0.0295 & 0.6241±0.0321          & 0.6083±0.0412          & 0.6253±0.0228 \\
			vowel.csv               & 0.4305±0.1500 & 0.4359±0.1575          & 0.4177±0.1596          & 0.4345±0.1651 & 0.3468±0.1489 & \textbf{0.4359±0.1575} & 0.3982±0.1505          & 0.3977±0.1670 \\
			\hline
		\end{tabular}
	}
\end{table*}


In this section, we compare our algorithm with other seven widely-used or state-of-the-art attribute selection algorithms on ten benchmark datasets randomly selected from the UCI machine learning library. Relevant metadata of these data sets are summarized in Table \ref{tab:DataInfo}. We choose seven baseline algorithms, including two rough set attribute selection algorithms and six other attribute selection algorithms. The former two algorithms include the classical rough set algorithm and its acceleration version using Theorem 1. Only the classical one is used for comparison in accuracy because both of them have the same attribute selection results. The latter six algorithms include CFS \cite{Guyon2002}, Fisher \cite{Gu2012}, ILFS \cite{Roffo2017a, Roffo2020}, Laplacian \cite{He2005}, LASSO \cite{Van2008}, and Mutinffs. All these six algorithms use the implementation provided in the Feature Selection Code Library (FSLib) \cite{Roffo2020} (https://www.mathworks.cn/matlabcentral/fileexchange/56937-feature-selection-library). Regarding all eight algorithms for comparison, 10-fold cross-validation is used to evaluate the reduction results, and the average of the five classification accuracies is used as the final result.

\subsection{Effectiveness of RSLRS in attribute selections \label{subsec:EffectivenessAS}}

\begin{table}[htp]
	\renewcommand{\arraystretch}{1.3}
	\caption{Dataset Information}
	\label{tab:DataInfo}
	\centering
	\begin{tabular}{lll}
		\hline
		Data sets                     & Features & Samples \\
		\hline
		anneal                      & 21       & 798     \\
		Healthy\_Older\_People2.csv & 9        & 22646   \\
		hepatitis.csv               & 20       & 155     \\
		htru2                       & 9        & 17898   \\
		lymphography.csv            & 19       & 148     \\
		zoo.csv                     & 17       & 101     \\
		letter                      & 17       & 20000   \\
		GINA                        & 971      & 3153    \\
		ionosphere.csv              & 33       & 351     \\
		sensorReadings              & 25       & 5456   \\
		\hline
	\end{tabular}
\end{table}

The experimental results of different attribute selection methods in accuracy are shown in Table \ref{tab:FSAccurcy}. It can be observed that the RSLRS can provide a higher accuracy than other methods on most data sets, especially when comparing to CRS, because the model of relative importance evaluation can alleviate the overfitting issue in a rough set. Therefore, the RSLRS presents a higher generalizability than other rough set methods. 

\subsection{Effectiveness of rough set classifier and knowledge representation \label{subsec:EffectivenessClassifier}}

\begin{figure*}[htb]
	\centering
	\includegraphics[width = 0.8\textwidth]{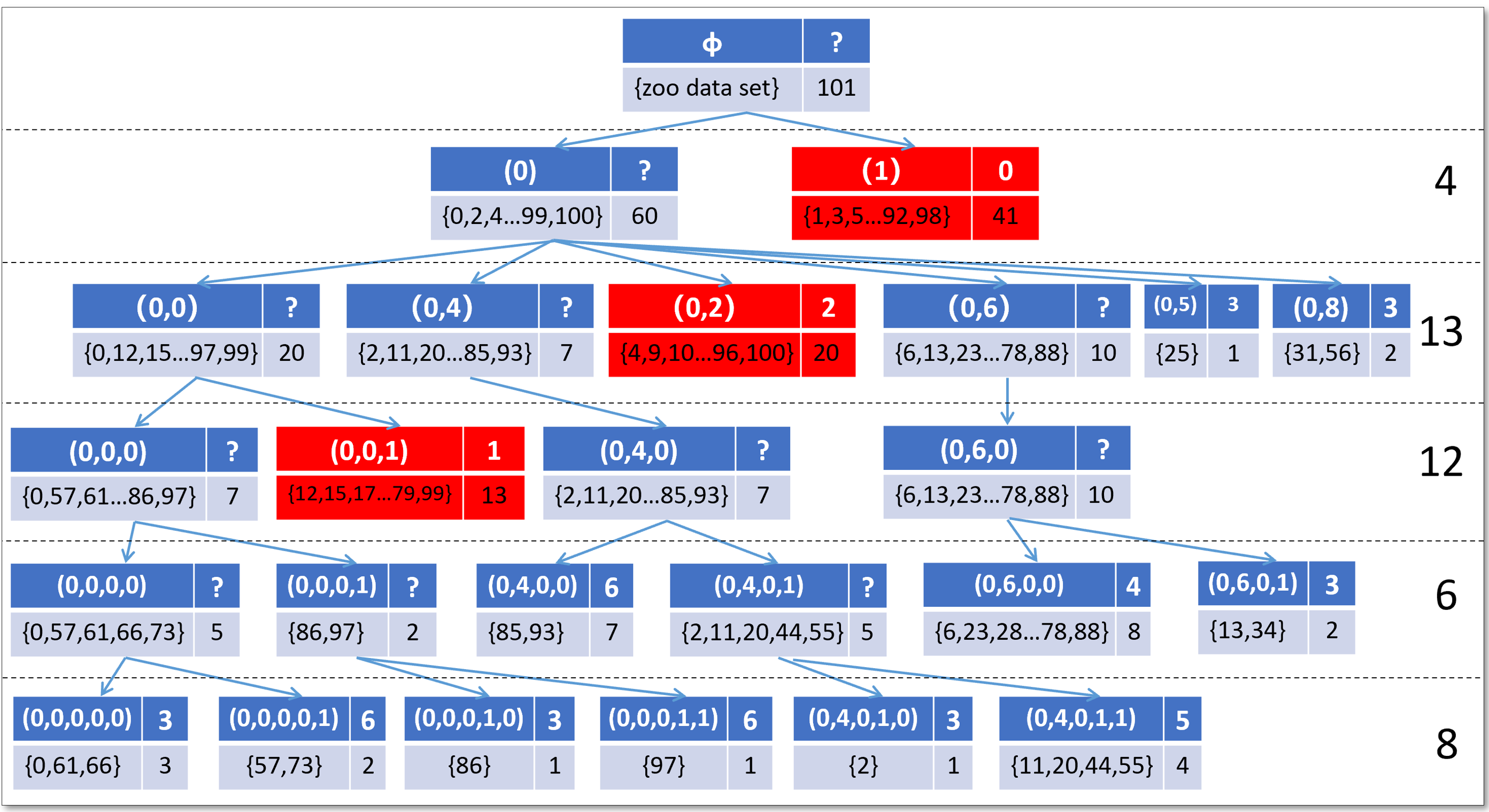}
	\caption{Knowledge repretation of the data set zoo using rough concept tree.}
	\label{fig:zooRCT}
\end{figure*}

Table \ref{tab:ClassifierAccuracy} presents the accuracies of each method. It can be observed that the RSC has a higher accuracy than others, including LR, on approximately half of the data sets. The knowledge nodes in the RCT consisting of both "intent" and "extent", as mentioned above, are very useful in many fields. Using the data set zoo as an example, there are sixteen conditional attributes including hair, feathers, eggs, milk, air, aquatic, predator, teeth, backbone, breath, venom, fins, legs, tail, whether it is domestic or not and catsize. The decision attribution of the zoo data set is the animal type such as mammals, fish, birds, invertebrates, insects, amphibians and reptiles, whose corresponding values range from 0 to 6. The RCT on the zoo is shown in Fig. \ref{fig:zooRCT}. In the RCT, all the knowledge nodes without a question mark can describe the certain knowledge to determine the animal type. Among them, the three knowledge nodes in red contain the most objects and have the largest knowledge description ability. For a new data point of an animal, if its third conditional attribute, i.e., milk, is equal to 1, we can use this knowledge point to assign it to the first animal type, i.e., mammals. In summary, the animals that produce milk are all mammals. Similarly, the other two red knowledge nodes indicate that an animal which does not produce milk with 2 legs is a bird; an animal with no legs but with fins which does not produce milk is a fish. These statements are easy to understand and to validate from a biological point of view.

\begin{table}[htp]
	\renewcommand{\arraystretch}{1.3}
	\caption{Accuracy comparison between rough set classifier and logic regression}
	\label{tab:ClassifierAccuracy}
	\centering
	\resizebox{.65\columnwidth}{!}{
		\begin{tabular}{lll}
			\hline
			& LR     & RSC \\
			\hline
			anneal                  & 0.8609 & 0.8271          \\
			Healthy\_Older\_People2 & 0.9561 & 0.9285          \\
			hepatitis               & 0.7867 & 0.7221          \\
			htru2                   & 0.9759 & 0.9620          \\
			lymphography            & 0.7961 & 0.7167          \\
			zoo                     & 0.8833 & \textbf{0.9409} \\
			letter                  & 0.6539 & \textbf{0.8362}     \\
			GINA                    & 0.7380 & 0.6492          \\
			ionosphere              & 0.7813 & \textbf{0.8576} \\
			sensorReadings          & 0.6253 & \textbf{0.8198} \\
			vowel                   & 0.3977 & \textbf{0.5622} \\
			ave                     & 0.7687 &
			\textbf{0.8020}\\
			\hline
		\end{tabular}
	}
\end{table}

\subsection{Efficiency of RSLRS \label{subsec:Efficiency}}

In this section, we compare the efficiency among multiple popular rough set algorithms. For a fair comparison in efficiency between different acceleration strategies in the rough set, the relative importance model is used in all the three rough set algorithms so that all of them have the same reduction results. The efficiency results of the experiment are shown in Fig. 4. Logarithm of running time is used to make all the results easier to visualize. It can be observed that the efficiency of the RSLRS is significantly improved in comparison with other acceleration methods for rough sets because our RSLRS framework can considerably decrease the number of equivalence classes calculation in computation for positive region using the stability of local redundancy.

\begin{figure}[ht]
	\centering
	\includegraphics[width = 0.5\textwidth]{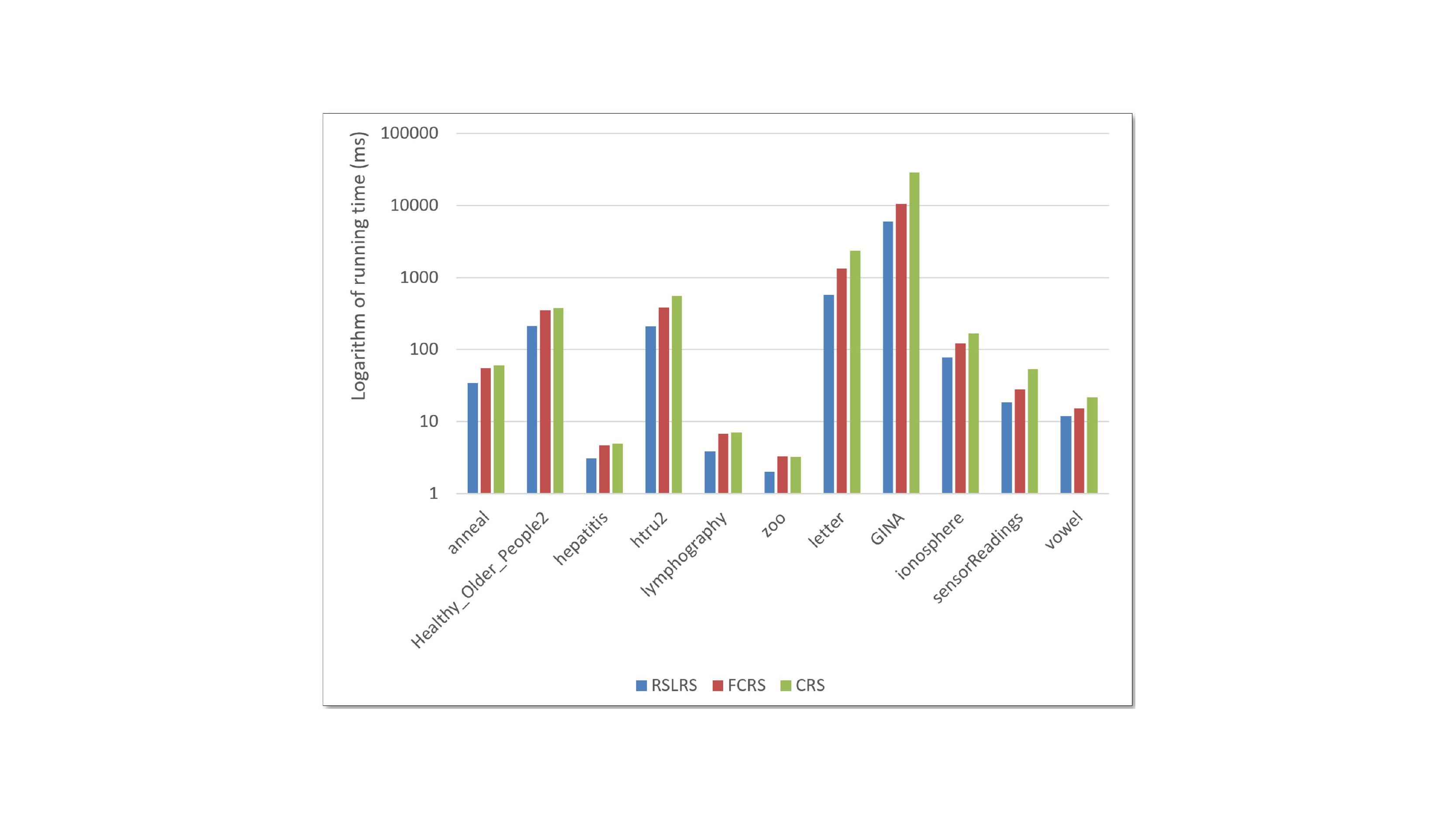}
	\caption{Efficiency comparison}
	\label{fig:Fig4}
	\vspace{-0.5cm}
\end{figure}

\section{Conclusion\label{Conclusion}}


This paper presents an effective framework to accelerate the rough set and improve its accuracy by proposing the theorem regarding the stability of the local redundancy of attributes and the model of relative importance. Besides, we develop the rough set to be a classifier and knowledge presentation tool. To the best of our knowledge, this work has enabled the rough set to be the only data mining method inherently effective in feature selection, classification, and knowledge representation. The experimental results validate that the proposed method can achieve a higher accuracy compared with the state-of-the-art attribute selection algorithms in most cases. We also show that our method is always more efficient than other acceleration rough set methods. This work significantly improves the ability of rough sets in feature selections when being applied in real-world scenarios. Future work will focus on further improving its efficiency for large-scale input data by parallelizing the current framework.

%

\section*{Acknowledgment}

Thanks to Ling Wei and Jinhai Li for their discussion about the concept lattice. This work was supported in part by National Key Research and Development Program of China (2019QY(Y)0301, the National Natural Science Foundation of China under Grant Nos. 62176033 and 61936001, the Natural Science Foundation of Chongqing No. cstc2019jcyj-cxttX0002. and NICE: NRT for Integrated Computational Entomology, US NSF award 1631776.


\begin{IEEEbiography}[{\includegraphics[width=1in,height=1.25in,clip,keepaspectratio]{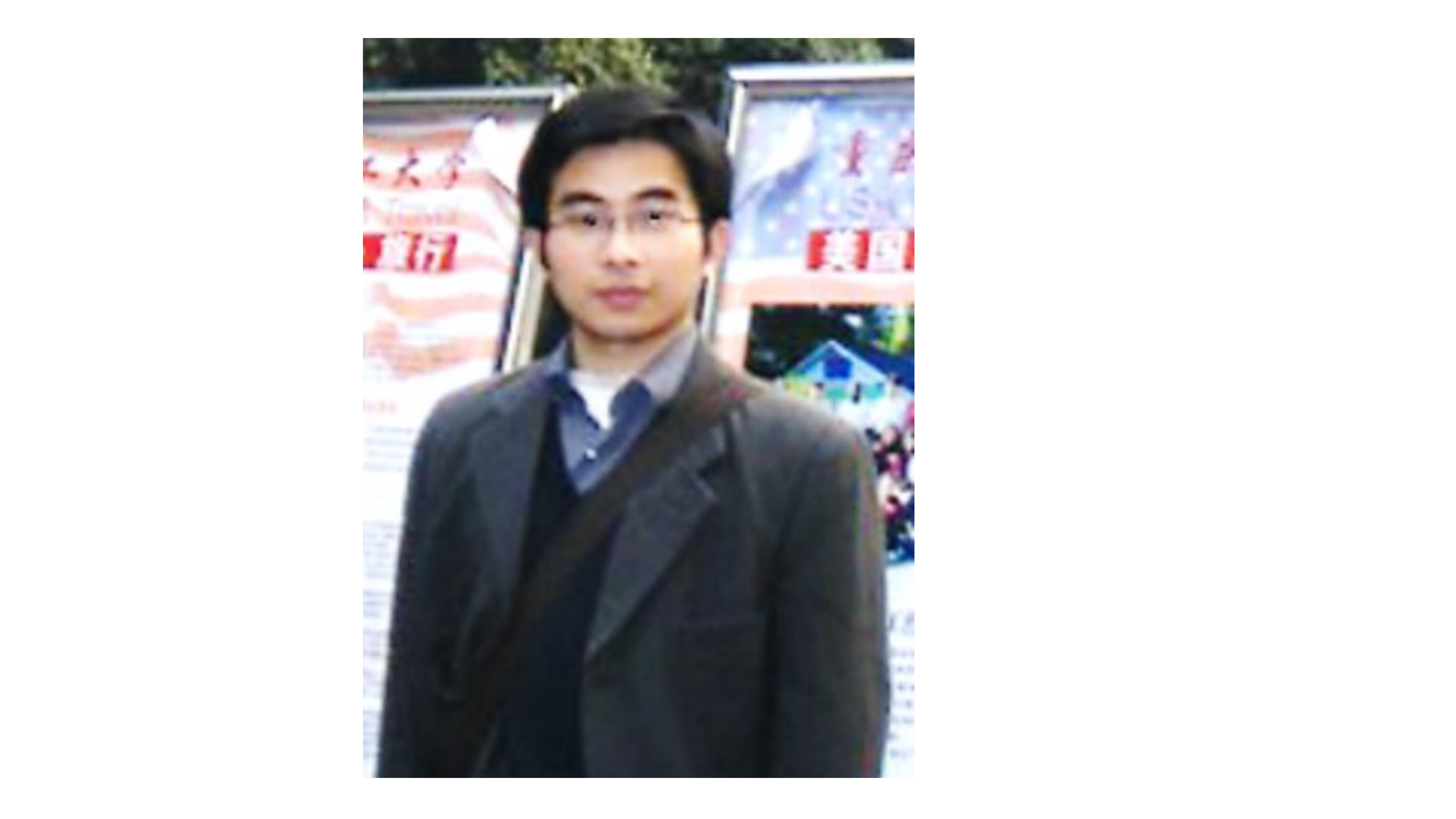}}]{Shuyin Xia}
	received his B.S. degree in 2008 and his M.S. degree in 2012, both in computer science and both from the Chongqing University of Technology in China. He received his Ph.D. degree from the College of Computer Science at Chongqing University in China. He is an associate professor and a Ph.D. supervisor at the Chongqing University of Posts and Telecommunications in Chongqing, China. His research interests include classifiers and granular computing. He has published more than 30 papers in prestigious journals and conferences, such as IEEE-TPAMI, IEEE-TKDE, IEEE-TNNLS and IS. He is the executive deputy director of the Big Data and Network Security Joint Lab of CQUPT and an IEEE Member.
\end{IEEEbiography}

\begin{IEEEbiography}[{\includegraphics[width=1in,height=1.25in,clip,keepaspectratio]{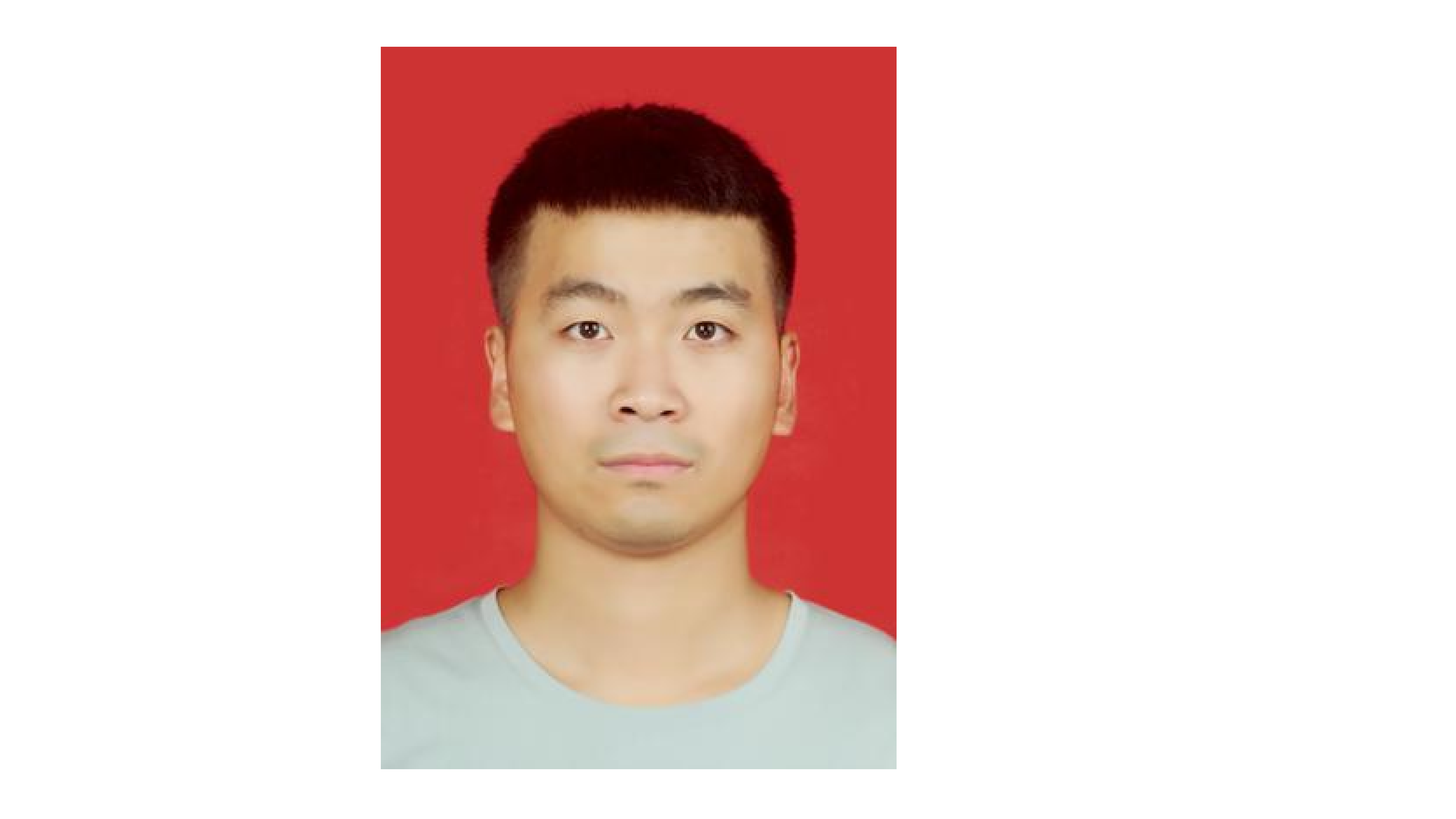}}]{Xinyu Bai}
	received her B.S. degree in Information and Computation Science in 2016, from Chongqing University of Posts and Telecommunications in Chongqing, China. He is currently pursuing the M.S.  degree in Computer Science and Technology in Chongqing University of Posts and Telecommunications in Chongqing, China. Her research interests include Rough sets, granular computing and data mining.
\end{IEEEbiography}


\begin{IEEEbiography}[{\includegraphics[width=1in,height=1.25in,clip,keepaspectratio]{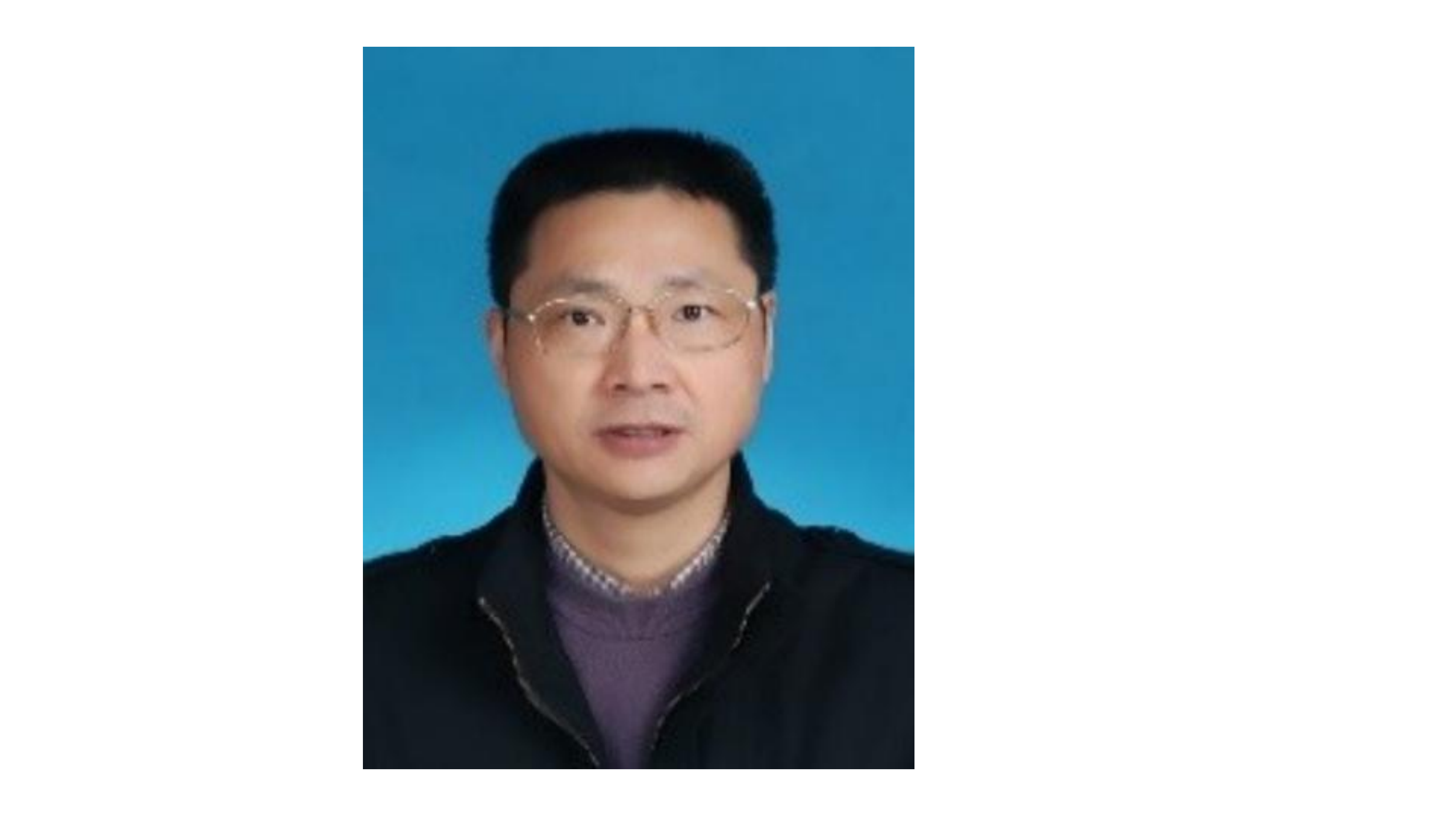}}]{Guoyin Wang}
	received a B.E. degree in computer software in 1992, a M.S. degree in computer software in 1994, and a Ph.D. degree in computer organization and architecture in 1996, all from Xi'an Jiaotong University in Xi'an, China. His research interests include data mining, machine learning, rough sets, granular computing, cognitive computing, and so forth. He has published over 300 papers in prestigious journals and conferences, including IEEE T-PAMI, T-KDE, T-IP, T-NNLS, and TCYB. He has worked at the University of North Texas, USA, and the University of Regina, Canada, as a Visiting Scholar. Since 1996, he has been working at the Chongqing University of Posts and Telecommunications in Chongqing, China, where he is currently a Professor and a Ph.D. supervisor, the Director of the Chongqing Key Laboratory of Computational Intelligence, and the Vice President of Chongqing University of Posts and Telecommunications. He is the Steering Committee Chair of the International Rough Set Society (IRSS), a Vice-President of the Chinese Association for Artificial Intelligence (CAAI), and a council member of the China Computer Federation (CCF).
\end{IEEEbiography}

\begin{IEEEbiography}[{\includegraphics[width=1in,height=1.25in,clip,keepaspectratio]{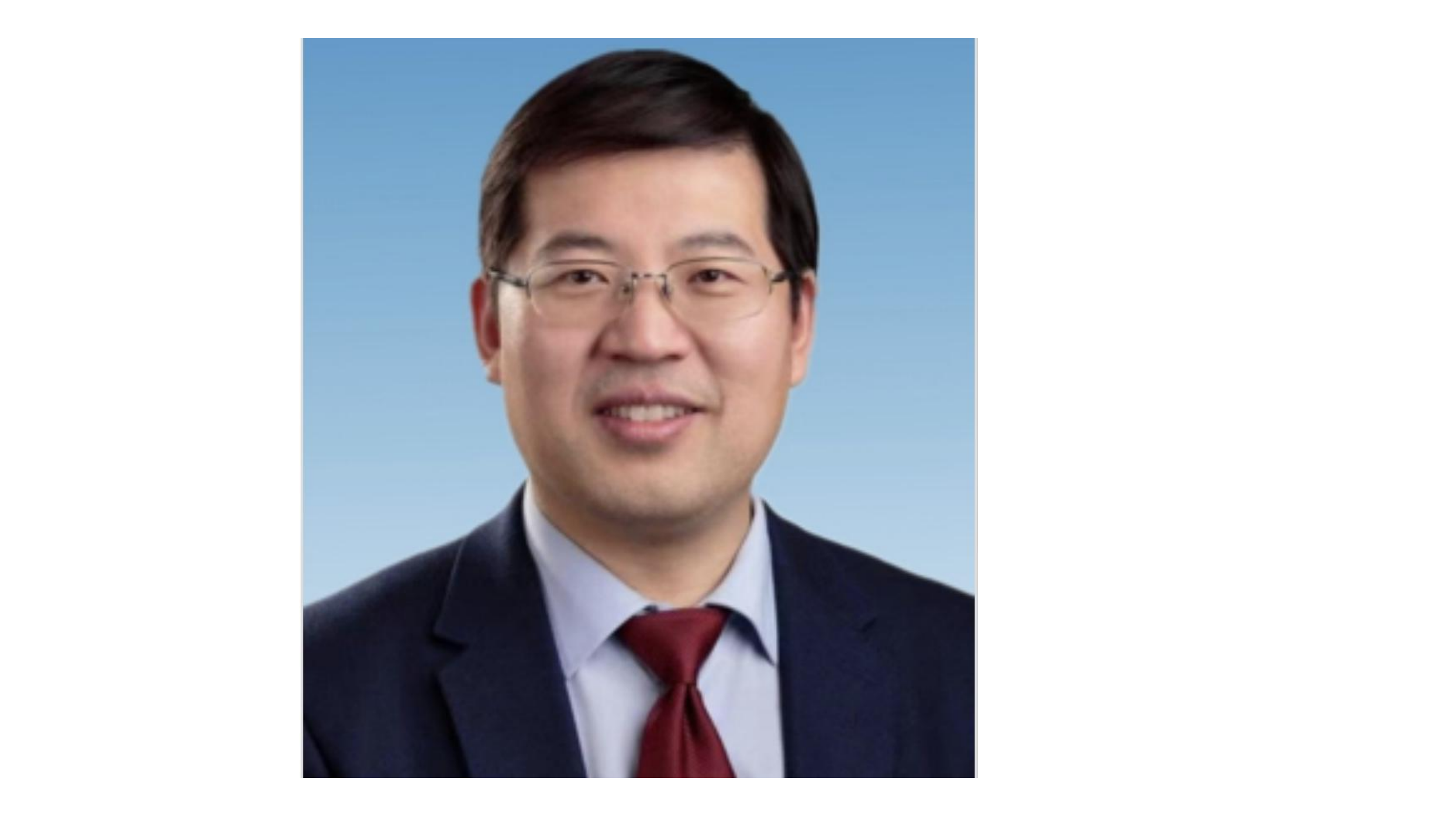}}]{Xinbo Gao}
	(M’02-SM’07) received the BEng, MSc, and PhD degrees in signal and information processing from Xidian University, Xi’an, China, in 1994, 1997, and 1999, respectively. From 1997 to 1998, he was a research fellow with the Department of Computer Science, Shizuoka University, Shizuoka, Japan. From 2000 to 2001, he was a post-doctoral research fellow with the Department of Information Engineering, the Chinese University of Hong Kong, Hong Kong. From 2001 to 2020, he was at the School of Electronic Engineering, Xidian University. He is currently the president of Chongqing University of Posts and Telecommunications. He has published six books and around 200 technical articles on prestigious journals and conferences, including IEEE T-PAMI, T-IP, T-NNLS, T-MI, NIPS, CVPR, ICCV, AAAI and IJCAI.
\end{IEEEbiography}

\begin{IEEEbiography}[{\includegraphics[width=1in,height=1.25in,clip,keepaspectratio]{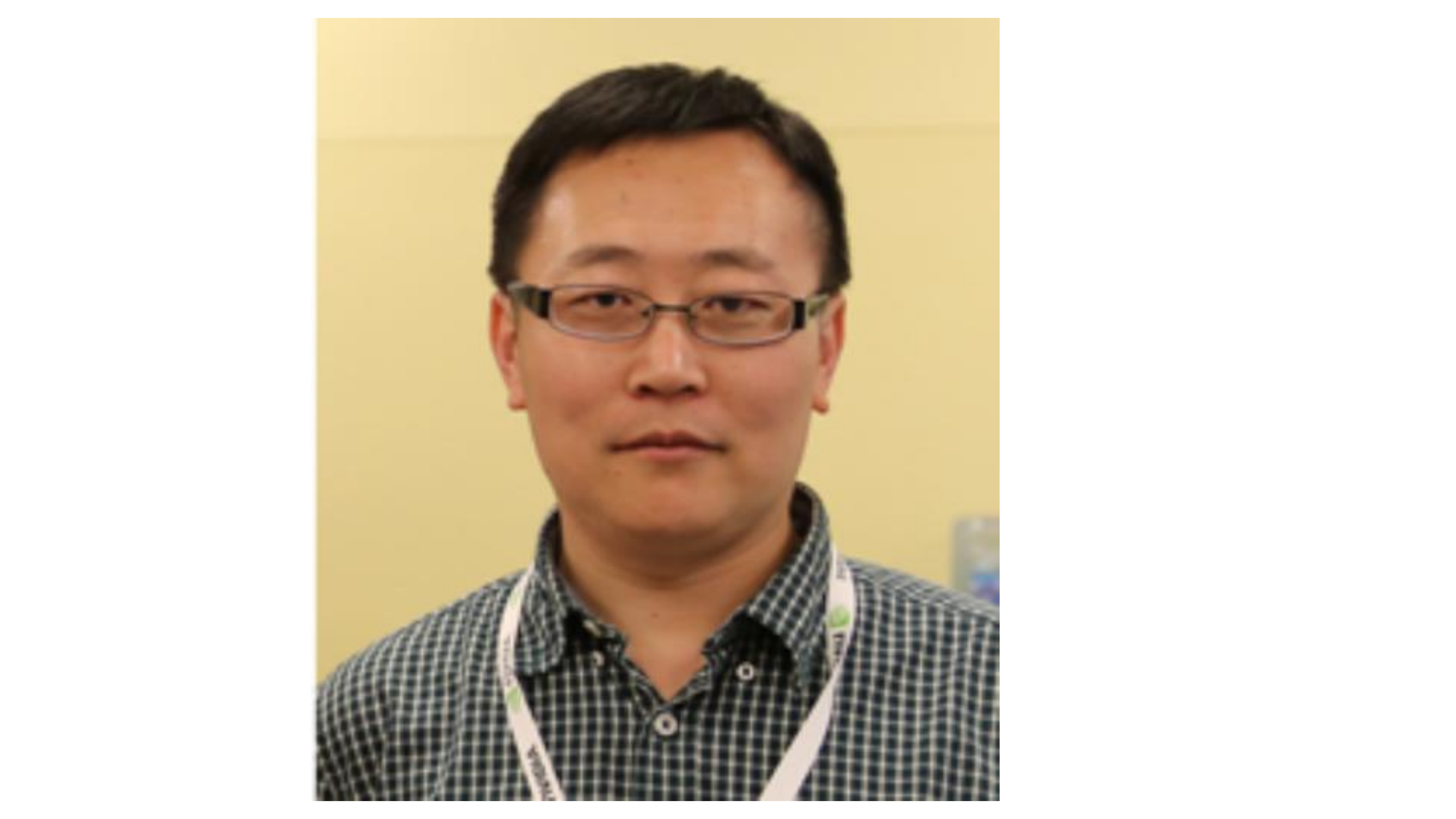}}]{Deyu Meng} received his B.Sc., M.Sc., and Ph.D. degrees from Xi'an Jiaotong University in Xi'an, China, where he is currently a Professor with the Institute for Information and System Sciences. His current research interests include self-paced learning, noise modeling, and tensor sparsity. He was a Visiting Scholar at Carnegie Mellon University in Pittsburgh, PA, USA. He is a regular Reviewer of ICML, NIPS, CVPR, ICCV, ECCV, AAAI, IJCAI, ACM MM, AISTATS, ACCV, ACML, ICPR, and IEEE T-PAMI, IEEE T-IP, IEEE T-NNLS and others. He is the Associate Editor of IEEE T-PAMI.
\end{IEEEbiography}

\begin{IEEEbiography}[{\includegraphics[width=1in,height=1.25in,clip,keepaspectratio]{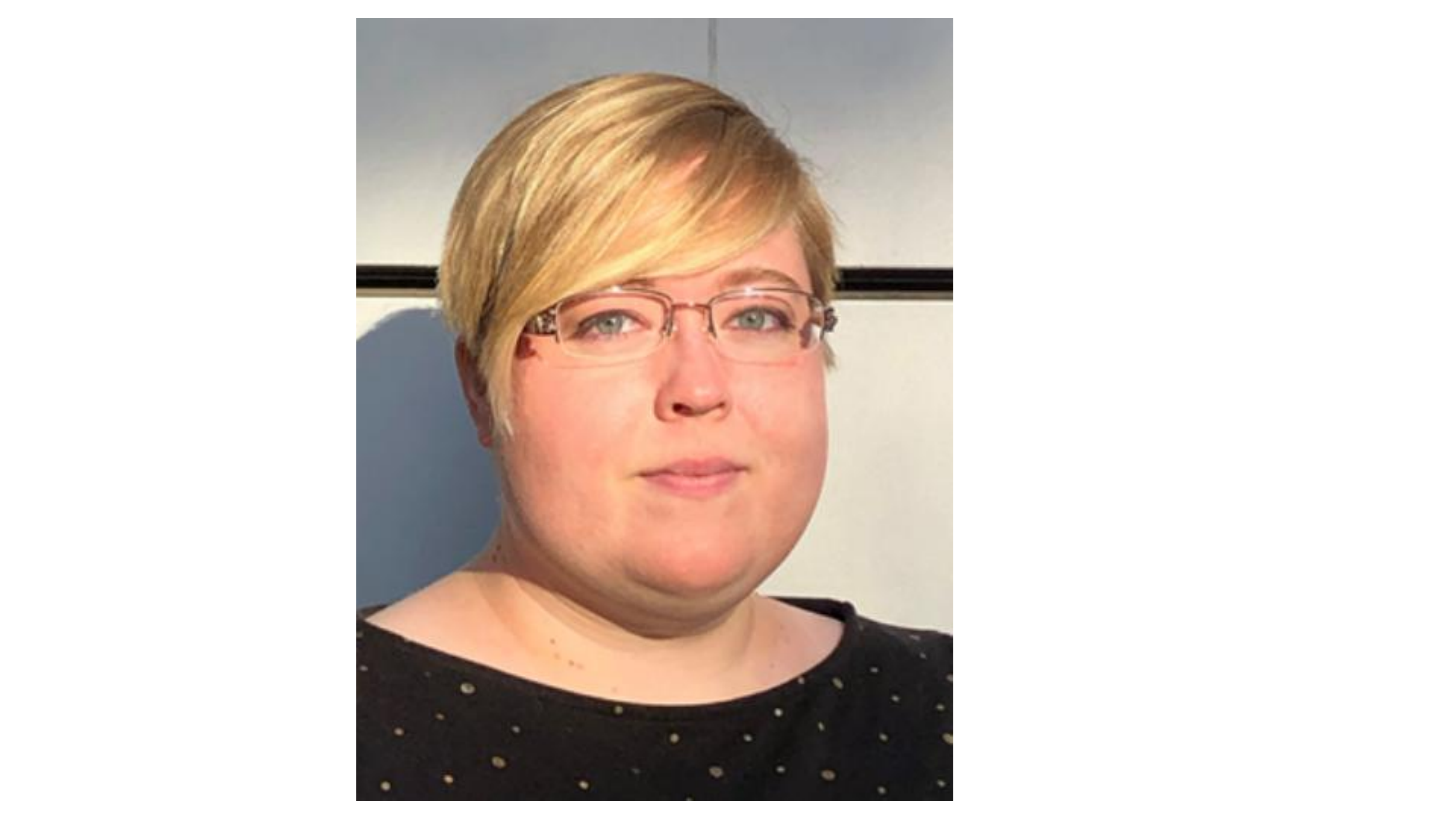}}]{Elisabeth Giem}
	received two bachelor's degrees, one in pure mathematics and one in music (concentration in performance) from the University of California, Riverside (UCR). She received a master's degree in computational and applied mathematics from Rice University, and a master's degree in pure mathematics from UCR. She joined Zizhong Chen's SuperLab in 2018 as a computer science Ph.D. student, and has been awarded the NSF NRT In Computational Entomology Fellowship. Her research interests include but are not limited to high-performance computing, parallel and distributed systems, big data analytics, computational entomology, and numerical linear algebra algorithms and software. He has published many papers in prestigious journals and conferences, such as IEEE-TPAMI,
	and IEEE-TKDE.
\end{IEEEbiography}

\end{document}